\documentclass[11pt]{article}

\usepackage[margin=1in]{geometry}

\usepackage[utf8]{inputenc}
\usepackage[T1]{fontenc}
\usepackage[english]{babel}
\usepackage{mathptmx} 

\usepackage{amsmath}
\usepackage{amssymb}
\usepackage{amsfonts}

\usepackage{graphicx}
\usepackage{subcaption}
\usepackage{booktabs}
\usepackage{multirow}
\usepackage{array}

\usepackage{color}
\usepackage{xcolor}
\usepackage{hyperref}
\hypersetup{
    colorlinks=true,
    linkcolor=blue,
    citecolor=blue,
    urlcolor=blue
}

\usepackage{float}
\usepackage{placeins}

\usepackage{cite}
\usepackage{url}
\usepackage{xurl}
\usepackage{algorithmic}
\usepackage{textcomp}

\usepackage{titlesec}
\usepackage{authblk}

\def\BibTeX{{\rm B\kern-.05em{\sc i\kern-.025em b}\kern-.08em
    T\kern-.1667em\lower.7ex\hbox{E}\kern-.125emX}}

\begin{document}

\title{Unveiling and Bridging the Functional Perception Gap in MLLMs: Atomic Visual Alignment and Hierarchical Evaluation via PET-Bench}

\author[1,*]{Zanting Ye}
\author[1,*]{Xiaolong Niu}
\author[1]{Xuanbin Wu}
\author[2]{Xu Han}
\author[3]{Shengyuan Liu}
\author[4]{Jing Hao}
\author[3]{Zhihao Peng}
\author[1]{Hao Sun}
\author[5]{Jieqin Lv}
\author[6]{Fanghu Wang}
\author[7]{Yanchao Huang}
\author[7]{Hubing Wu}
\author[3]{Yixuan Yuan}
\author[8]{Habib Zaidi}
\author[9]{Arman Rahmim}
\author[10,$\dagger$]{Yefeng Zheng}
\author[1,$\dagger$]{Lijun Lu}

\affil[1]{School of Biomedical Engineering, Southern Medical University, Guangzhou, China}
\affil[2]{School of Biomedical Engineering, Shanghai Jiaotong University, Shanghai, China}
\affil[3]{Department of Electronic Engineering, Chinese University of Hong Kong, Hong Kong, China}
\affil[4]{Faculty of Dentistry, The University of Hong Kong, Hong Kong, China}
\affil[5]{Department of Nuclear Medicine, The Second Affiliated Hospital of Guangzhou University of Chinese Medicine, Guangzhou, China}
\affil[6]{PET Center, Department of Nuclear Medicine, Guangdong Provincial People's Hospital, Southern Medical University, Guangzhou, China}
\affil[7]{Department of Nuclear Medicine, Nanfang Hospital, Southern Medical University, Guangzhou, China}
\affil[8]{Division of Nuclear Medicine and Molecular Imaging, Geneva University Hospitals, Geneva, Switzerland}
\affil[9]{Departments of Radiology, Physics, and Biomedical Engineering, The University of British Columbia, Vancouver, Canada}
\affil[10]{Medical Artificial Intelligence Laboratory, Westlake University, Hangzhou, China}

\affil[*]{Equal contribution}
\affil[$\dagger$]{Corresponding authors: zhengyefeng@westlake.edu.cn; ljlubme@gmail.com}

\date{}

\maketitle

\begin{abstract}
While Multimodal Large Language Models (MLLMs) have demonstrated remarkable proficiency in tasks such as abnormality detection and report generation for anatomical modalities, their capability in functional imaging remains largely unexplored. In this work, we identify and quantify a fundamental functional perception gap: the inability of current vision encoders to decode functional tracer biodistribution independent of morphological priors. Identifying Positron Emission Tomography (PET) as the quintessential modality to investigate this disconnect, we introduce PET-Bench, the first large-scale PET benchmark comprising 52,308 hierarchical QA pairs from 9,732 multi-site, multi-tracer PET studies. Extensive evaluation of 19 state-of-the-art MLLMs reveals a critical safety hazard termed the Chain-of-Thought (CoT) hallucination trap. We observe that standard CoT prompting, widely considered to enhance reasoning, paradoxically decouples linguistic generation from visual evidence in PET, producing clinically fluent but factually ungrounded diagnoses. To resolve this, we propose Atomic Visual Alignment (AVA), a simple fine-tuning strategy that enforces the mastery of low-level functional perception prior to high-level diagnostic reasoning. Our results demonstrate that AVA effectively bridges the perception gap, transforming CoT from a source of hallucination into a robust inference tool and improving diagnostic accuracy by up to 14.83\%. Code and data are available at \url{https://github.com/yezanting/PET-Bench}.
\end{abstract}


\begin{figure}[!htbp]
\centering
\includegraphics[width=\linewidth]{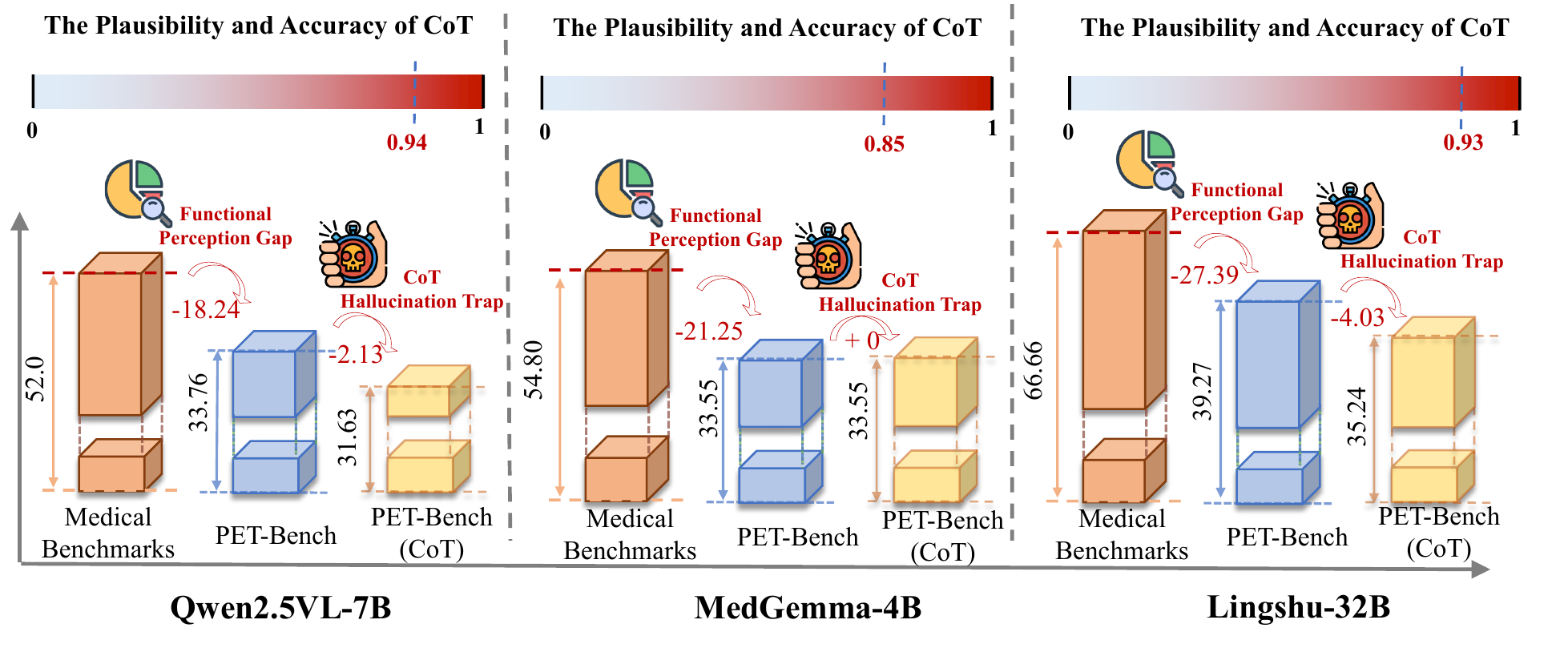}
\caption{Illustration of the two critical failure modes identified in MLLMs when applied to functional imaging. Functional Perception Gap: While current SOTA models achieve high performance on structural imaging benchmarks (the results from Lingshu test data), their zero-shot diagnostic accuracy on PET drops significantly. The CoT Hallucination Trap: Attempting to bridge this gap via standard CoT prompting paradoxically degrades reliability. Without domain-specific visual grounding, models generate linguistically plausible but factually ungrounded rationales, leading to high-confidence misdiagnoses.}
\label{fig:motivation}
\end{figure}

\section{Introduction}
\label{sec:introduction}

Multimodal Large Language Models (MLLMs) have recently catalyzed a paradigm shift in medical artificial intelligence, demonstrating robust capabilities in interpreting anatomical modalities such as radiography, Computed Tomography (CT), and Magnetic Resonance Imaging (MRI) \cite{xu2025lingshu,li2024gmai,sellergren2025medgemma,jiang2025hulu}. By integrating advanced visual encoders with large-scale language reasoning, state-of-the-art models like GPT-5, Gemini-2.5-Pro \cite{comanici2025gemini} and domain-specific adaptations \cite{sellergren2025medgemma,xu2025lingshu} have shown remarkable success in abnormality detection and report generation. These advances have catalyzed expectations for AI-assisted diagnosis and clinical decision support in radiology \cite{goh2025gpt,yao2025multimodal,liu2025generalist}. Despite these advances, a critical dichotomy remains: while current methodologies excel at characterizing structural morphology (e.g., tissue density and proton relaxation), their capability to interpret functional imaging remains largely unexplored and unverified \cite{xu2025lingshu,zhu2025well}. This limitation is particularly acute in nuclear medicine, where diagnosis necessitates the quantification of dynamic molecular physiology and functional and molecular kinetics rather than static anatomical boundaries.

Positron Emission Tomography (PET) constitutes a fundamental metabolic and molecular imaging modality in oncology, neurology, and cardiology. Unlike CT and MRI, which provide high-frequency structural details, PET visualizes the biodistribution of radiotracers as a surrogate for underlying functional activity. Interpreting these low spatial resolution functional and molecular heatmaps requires reasoning about tracer-specific uptake intensity, distinguishing physiological background from pathological hypermetabolism, and accounting for low-count noise. These functional semantics often poorly align with the high-frequency edge and texture features prioritized by standard vision encoders pretrained on natural images or structural medical datasets.

From both data coverage and model-training perspectives, PET is substantially under-represented in the current MLLM ecosystem. Existing medical MLLMs are typically derived by adapting general-domain MLLMs via instruction tuning on radiology reports, medical textbooks, and large collections of X-ray, CT, and MRI images \cite{xu2025lingshu,zhang2023huatuogpt}. As summarized in Table~\ref{tab:medical_vlm_summary}, public specifications of many medical MLLMs do not explicitly include PET as a supervised modality. Due to the relatively high cost of PET imaging, available PET resources are limited in scale, often restricted to FDG studies, and primarily target segmentation or free-text reporting without structured labels for image quality, organ-level uptake, or intermediate perceptual tasks (Table~\ref{tab:dataset_comparison}). This structural bias motivates our hypothesis of a functional perception gap: a fundamental deficit where MLLMs, conditioned heavily on anatomical features, lose the capability to represent tracer biodistribution and voxel intensity independent of morphological priors. Consequently, these models fail to quantify dynamic metabolic and molecular processes, instead hallucinating plausible anatomical descriptions that are ungrounded in the functional signal. This hypothesis is empirically substantiated in Fig.~\ref{fig:motivation}, where we observe a precipitous performance drop when shifting models from structural to functional tasks, confirming that current architectures struggle to generalize beyond anatomical recognition to true functional interpretation.

Chain-of-Thought prompting \cite{wei2022chain} has emerged as a widely adopted strategy for eliciting multi-step reasoning in large language models across mathematical, logical, and clinical decision-making tasks \cite{wang2025v2t,kim2025small}. By encouraging models to articulate intermediate steps, CoT often enhances performance and interpretability in purely textual and structurally grounded multimodal settings. However, most existing evidence implicitly assumes that the underlying visual perception is accurate, enabling CoT to operate on reliable intermediate representations \cite{barez2025chain,mayne2025llms}. In functional imaging, this assumption is fragile. When visual grounding is weak, CoT may inadvertently amplify language priors, producing reasoning trajectories that appear clinically plausible but lack support from the image data. Whether CoT genuinely improves PET-based diagnosis or primarily increases the risk of confident but visually ungrounded explanations remains an open and clinically critical question.

In this work, we systematically characterize both the functional perception gap and the behavior of CoT-based reasoning for PET within a unified evaluation framework. We introduce PET-Bench, a multi-site, multi-tracer benchmark specifically designed for functional imaging. PET-Bench is constructed from Standardized Uptake Value (SUV)-normalized pure PET volumes without PET/CT overlays, thereby isolating functional understanding from structural priors. Guided by the cognitive workflow of nuclear medicine physicians, we structure the benchmark into a five-level hierarchy, progressing from global image perception (tracer identification and image quality assessment) to semantic grounding (organ recognition and abnormality detection), and finally to disease diagnosis. This hierarchical taxonomy allows us to pinpoint exactly where the reasoning chain breaks down: whether the model fails to see the lesion (perception gap) or fails to interpret it (reasoning deficit).

Using PET-Bench, we perform a comprehensive evaluation of a diverse panel of state-of-the-art MLLMs. We further probe diagnostic reasoning with a standardized, six-step CoT prompt mirroring clinical workflows. Our analysis reveals that current MLLMs exhibit pronounced, task-dependent deficits on PET, and that linguistically coherent CoT explanations do not guarantee correct, molecularly and grounded decisions. We term this phenomenon the CoT hallucination trap: a failure mode where the generated reasoning chain maintains high linguistic plausibility yet diverges significantly from the underlying functional signals, as illustrated in Fig.~\ref{fig:motivation}.

To address these limitations, we propose a simple Atomic Visual Alignment strategy that decomposes PET interpretation into a set of clinically meaningful, atomic visual tasks used to explicitly align model representations. Concretely, we perform supervised fine-tuning on underlying visual understanding tasks in PET-Bench, enforcing that models acquire robust tracer- and organ-level metabolic and molecular perception before addressing diagnosis. To prevent data leakage, we enforce a strict patient-level split: the test set for Level 5 diagnosis consists exclusively of patients who were never exposed to the model during the AVA training phase (Levels 1–4). This ensures that the evaluation reflects true diagnostic generalization rather than the memorization of patient-specific physiological distributions. This alignment is conceptually distinct from generic instruction tuning: it treats low- and mid-level PET perception as a prerequisite for high-level reasoning, rather than as incidental by-products of diagnostic supervision. Under this regime, CoT transitions from an unreliable intervention into a consistently beneficial reasoning mechanism for PET-based diagnosis.

In summary, this work makes three main contributions:
\begin{itemize}
    \item We introduce PET-Bench, the first large-scale PET-focused benchmark with a five-level hierarchical VQA design that explicitly decomposes PET interpretation from low-level functional perception to high-level diagnosis.
    \item We provide a systematic evaluation of a broad range of MLLMs on PET-Bench, identifying two critical phenomena in functional imaging: a persistent functional perception gap and a CoT hallucination trap, where fluent reasoning is not a reliable proxy for correctness.
    \item We introduce a simple yet effective AVA, a training paradigm that effectively bridges the perception gap. Our experiments demonstrate that AVA transforms CoT from a source of hallucination into a robust inference mechanism, significantly improving diagnostic accuracy and reliability.
\end{itemize}
These contributions establish PET-Bench as a principled testbed for functional imaging, highlight the limitations of directly transferring anatomical-image MLLMs to PET, and suggest a general methodological template for developing safer, visually grounded MLLMs for PET, SPECT, and other functional modalities.

\begin{table}[!htbp]
\centering
\small
\setlength{\tabcolsep}{3pt}
\caption{Summary of representative medical MLLMs and their training exposure to PET. ``$\checkmark$'' denotes explicit mention of PET (or PET/CT) in training data descriptions; ``$\times$'' indicates absence.}
\begin{tabular}{lll}
\toprule
\textbf{Medical MLLMs} & \textbf{Backbone Architecture} & \textbf{PET} \\
\midrule
ShizhenGPT \cite{chen2025shizhengpt}  & Qwen-2.5 (7B, 32B)              & $\times$ \\
HealthGPT \cite{lin2025healthgpt}   & CLIP-L/14                       & $\times$ \\
HuaTuoGPT \cite{zhang2023huatuogpt}  & LLaVA-1.5                       & $\times$ \\
MedGemma \cite{sellergren2025medgemma}    & Gemma 3 (4B, 27B) + SigLIP-400M & $\times$ \\
MedVLM-R1 \cite{pan2025medvlm}   & Qwen2-VL-2B                     & $\times$ \\
MedDr  \cite{he2024meddr}     & InternVL-40B                    & $\times$ \\
Med-R1 \cite{lai2025med}     & Qwen2-VL-2B                     & $\times$ \\
Lingshu \cite{xu2025lingshu}    & Qwen2.5-VL-7B/32B               & $\checkmark$ \\
\bottomrule
\end{tabular}
\label{tab:medical_vlm_summary}
\end{table}

\section{Related Work}
\label{sec:related_work}

\subsection{Multimodal Large Language Models for Medical Imaging}

MLLMs have achieved impressive results on a range of anatomical imaging tasks \cite{pai2025vision,shui2025large}. General-purpose architectures such as CLIP \cite{radford2021learning}, LLaVA \cite{liu2023visual}, and Qwen-VL \cite{bai2025qwen2} typically serve as backbones, adapted to the medical domain via instruction tuning or lightweight fine-tuning on image--report pairs and VQA datasets \cite{ye2025multimodal,lin2025taming}. To enhance domain-specific reasoning, specialized models including MedGemma \cite{sellergren2025medgemma}, Lingshu \cite{xu2025lingshu}, and ShizhenGPT \cite{chen2025shizhengpt} incorporate extensive training on radiology reports, medical textbooks, and large-scale collections of X-ray, CT, and MRI volumes.

Despite these advances, a critical modality gap persists. As summarized in Table~\ref{tab:medical_vlm_summary}, the supervision signal for these models is overwhelmingly dominated by structural imaging. Publicly available technical reports for leading medical MLLMs rarely list PET as an explicit training modality. A notable exception is Lingshu \cite{xu2025lingshu}, which reports the use of PET/CT fusion data. However, reliance on fused modalities presents a confound: models may learn to extract morphological features from the high-resolution CT component while ignoring the lower-resolution, metabolic and molecular PET signal. Consequently, the capability of current MLLMs to interpret pure functional data, where diagnosis depends on tracer uptake intensity and biodistribution rather than tissue density, remains unverified. PET-Bench addresses this by isolating the functional component, thereby rigorously testing the transferability of anatomical priors to metabolic and molecular imaging.

\subsection{Medical Imaging Benchmarks and PET Datasets}
The evaluation of medical MLLMs has been propelled by large-scale benchmarks \cite{lin2023pmc,ruckert2024rocov2,abacha2019vqa,liu2021slake,hao2025oralgpt, peng2025omnibrainbench}. Image--text corpora such as PMC-OA \cite{lin2023pmc} and ROCOv2 \cite{ruckert2024rocov2} facilitate generic captioning capabilities, while VQA benchmarks like VQA-Med \cite{abacha2019vqa} and SLAKE \cite{liu2021slake} assess natural language reasoning over radiographs and cross-sectional anatomy. However, these resources provide minimal coverage of functional imaging. Existing PET-specific datasets are generally constrained by scale, tracer diversity, and task formulation. As detailed in Table~\ref{tab:dataset_comparison}, datasets such as RIDER Lung PET-CT \cite{li2020large}, head--neck cohorts \cite{vallieresdata}, Lung-PET-CT-Dx \cite{li2020lungpetctdx}, and FDG-PET-CT-Lesions \cite{gatidis2022whole} typically contain fewer than 1,000 studies and are restricted to single-tracer FDG imaging. Furthermore, the primary annotations in these datasets are segmentation masks or findings extraction, which support discriminative tasks but do not probe high-level reasoning. The AutoPET challenge \cite{gatidis2024results} introduces dual-tracer data (FDG and PSMA) but remains focused on segmentation and quantitative volumetry. Recent efforts like PET2Rep \cite{zhang2025pet2rep} and ViMed-PET \cite{nguyen2025toward} target report generation; however, by evaluating holistic report quality without intermediate grounding checks, these benchmarks risk rewarding hallucinated narratives that mimic the style of radiology reports without accurately reflecting the specific image content.

\begin{table}[!t]
\centering
\small
\setlength{\tabcolsep}{3pt}
\caption{Comparison of representative medical imaging datasets and PET-Bench.}
\resizebox{\textwidth}{!}{
\begin{tabular}{llllll}
\toprule
\textbf{Dataset} & \textbf{3D Volume} & \textbf{Tracers} & \textbf{Quality Annot.} & \textbf{Scale} & \textbf{Task Type} \\
\midrule
PMC-OA \cite{lin2023pmc}             & $\times$ & --               & $\times$ & 600K  & Captioning    \\
ROCOv2 \cite{ruckert2024rocov2}          & $\times$ & --               & $\times$ & 432   & Captioning    \\
RIDER Lung \cite{li2020large}          & $\checkmark$ & FDG          & $\times$ & 274   & Findings      \\
Head-Neck \cite{vallieresdata}          & $\checkmark$ & FDG          & $\times$ & 504   & Findings      \\
Lung-PET-CT-Dx \cite{li2020lungpetctdx}     & $\checkmark$ & FDG          & $\times$ & 355   & Findings      \\
FDG-PET-CT-Lesions \cite{gatidis2022whole}  & $\checkmark$ & FDG          & $\times$ & 1,014 & Segmentation  \\
AutoPET III  \cite{gatidis2024results}        & $\checkmark$ & FDG, PSMA    & $\times$ & 1,204 & Segmentation  \\
PET2Rep \cite{zhang2025pet2rep}            & $\checkmark$ & FDG          & $\times$ & 565   & Report       \\
ViMed-PET  \cite{nguyen2025toward}         & $\checkmark$ & FDG          & $\times$ & 2,757 & Report   \\
\midrule
\textbf{PET-Bench (Ours)} 
                    & \textbf{$\checkmark$} 
                    & \textbf{FDG, PSMA, FAPI, MET} 
                    & \textbf{$\checkmark$} 
                    & \textbf{9,732} 
                    & \textbf{5-Level VQA}  \\
\bottomrule
\end{tabular}
}
\label{tab:dataset_comparison}
\end{table}

\subsection{Chain-of-Thought Reasoning and Visual Grounding}

CoT prompting \cite{wei2022chain} has become a standard paradigm for eliciting multi-step reasoning in LLMs, effectively linearizing complex inference tasks. In the medical domain, CoT has been adapted for clinical decision support and report generation, aiming to mirror the structured workflow of human experts \cite{le2025s,wu2024chain}. However, the efficacy of CoT in multimodal settings is predicated on the assumption of accurate visual perception. Prior work largely assumes that if the reasoning logic is sound, the diagnostic output will be correct \cite{barez2025chain,mayne2025llms}. This assumption breaks down in functional imaging, where the visual signal is subtle and physics-dependent. When an MLLM lacks the requisite features to interpret PET (the functional perception gap), CoT prompting can decouple the generated text from the image data, leading to fluent but factually incorrect explanations. We term this the CoT hallucination trap.

Our work aligns with recent efforts in hierarchical task decomposition \cite{zhou2025drvd} but applies it specifically to resolve this grounding failure in functional imaging. By enforcing AVA on the lower levels of the PET-Bench hierarchy, we demonstrate that CoT can be transformed from a source of hallucination into a robust, visually grounded diagnostic tool.

\begin{figure*}[!t]
\centering
\includegraphics[width=0.95\linewidth]{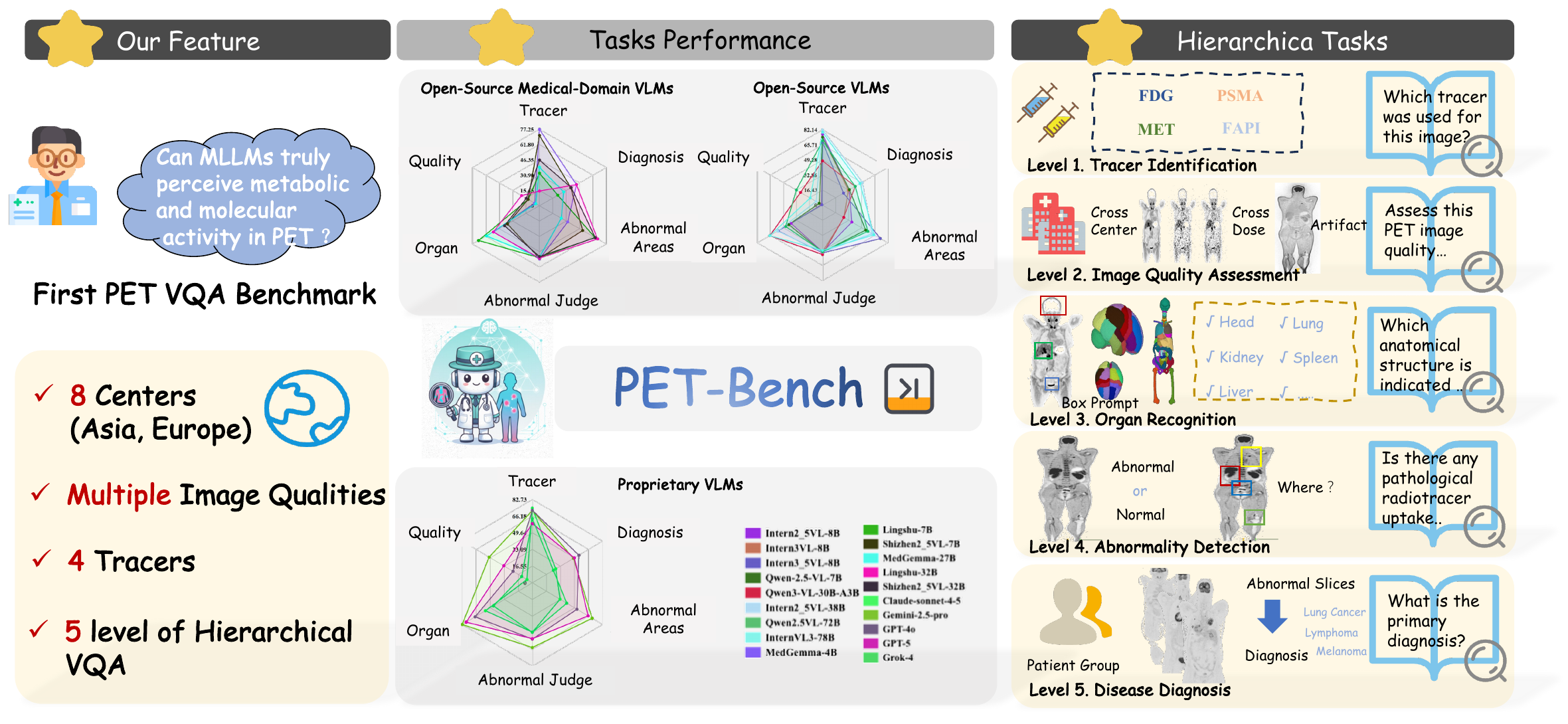}
\caption{Overview of the PET-Bench framework.
PET-Bench is the first large-scale benchmark designed to evaluate functional imaging capabilities, aggregating 52,308 QA pairs from 9,732 studies across 8 international centers. 
Unlike generic VQA datasets, the benchmark employs a five-level hierarchical taxonomy mirroring the nuclear medicine interpretation workflow: progressing from atomic perception to lesion detection, and finally to disease diagnosis. This structure allows for explicitly decoupling perceptual failures from reasoning deficits.}
\label{fig:overview}
\end{figure*}

\section{PET-Bench}
\label{sec:pet_bench}

To rigorously quantify the functional perception gap in current MLLMs, we introduce PET-Bench, a large-scale, multi-center benchmark designed to disentangle intrinsic functional perception from reliance on external anatomical priors. Fig. \ref{fig:overview} provides an overview of PET-Bench, highlighting its key features, dataset composition, and hierarchical task structure. Given that current medical MLLMs are predominantly pre-trained on large-scale structural imaging datasets, they inherently possess strong anatomical priors. Consequently, in existing benchmarks relying on PET/CT fusion, it remains ambiguous whether successful diagnosis stems from decoding the underlying radiotracer biodistribution or merely exploiting these entrenched high-resolution CT features. Prior studies have established that pure PET imaging retains sufficient spatial information for basic organ localization and structural grounding \cite{ye2025self,li2024learning}. Leveraging this property, PET-Bench explicitly utilizes pure PET data not to eliminate structural context, but to compel models to derive this context solely from metabolic signals rather than high-frequency CT overlays. This design serves as a rigorous stress test, ensuring that performance metrics reflect true metabolic interpretation capabilities rather than anatomical pattern matching. To our knowledge, this constitutes the largest PET cohort to date, spanning four distinct radiotracers. Uniquely, we integrate image quality assessment as a mandatory prerequisite, grounding downstream diagnostic reliability in verified signal integrity.

\begin{table*}[!t]
\centering
\small
\caption{Detailed Breakdown of the 8 Data Centers Comprising PET-Bench}
\label{tab:two_column_layout}
\setlength{\tabcolsep}{6pt}
\begin{tabular}{@{\extracolsep{\fill}}lllr | lllr}
\toprule
\textbf{ID} & \textbf{Center/Scanner} & \textbf{Details} & \textbf{Vol.} & \textbf{ID} & \textbf{Center/Scanner} & \textbf{Details} & \textbf{Vol.} \\
\midrule
1 & AutoPET (Various) & FDG/PSMA (Multi) & 1,611 & 5 & SMU Nanfang (uExpl.) & FDG/FAPI (Multi) & 3,003$^{\dagger}$ \\
2 & GHSG (Various) & FDG (Lymphoma) & 525 & 6 & SMU Nanfang (mCT) & FDG (Lung Cancer) & 310 \\
3 & U. Bern (Quadra) & FDG/PSMA (Multi) & 1,750$^{\dagger}$ & 7 & GPH-CM (Quadra) & MET/FDG (Myeloma) & 150 \\
4 & Ruijin (uExplorer) & FDG/PSMA (Multi) & 2,100$^{\dagger}$ & 8 & GPH-People (uExpl.) & FDG/PSMA (Multi) & 283 \\
\bottomrule
\multicolumn{6}{l}{\textit{$^{\dagger}$ Includes multi-dose reconstructions to simulate varying image qualities.}}
\end{tabular}
\end{table*}

\subsection{Data Curation and SUV Normalization}
PET-Bench aggregates 52,308 expert-curated question-answer pairs derived from 9,732 whole/total-body PET studies. Data were sourced from eight clinical centers across Asia and Europe, incorporating both public repositories (AutoPET III \cite{gatidis2024results}, Ultra-low Dose PET \cite{r69}) and four proprietary in-house cohorts. A distinguishing feature of our benchmark is the inclusion of data from Whole-Body PET/CT and state-of-the-art Total-Body PET/CT systems. The data acquisition covers a wide spectrum of image qualities, ranging from high-statistics clinical standard scans to simulated ultra-low-dose reconstructions. Detailed data information is available in Table~\ref{tab:two_column_layout}.

Raw PET data represent radioactivity concentration $C_{\mathrm{raw}} \in \mathbb{R}^{H \times W \times D}$ (Bq/mL), which is subject to high inter-scanner variance due to differing acquisition protocols. To mitigate distribution shifts in the input space of the MLLM, we project all volumes onto a physiologically standardized manifold using the SUV. Let $A_{\mathrm{inj}}$ denote the injected dose (Bq) at time $t_{\mathrm{inj}}$, and $t_{\mathrm{scan}}$ be the acquisition start time. The SUV-normalized volume $\text{SUV}$ is derived as:
\begin{equation}
\text{SUV} = \frac{C_{\mathrm{raw}} \cdot W}{A_{\mathrm{inj}} \cdot e^{-\lambda (t_{\mathrm{scan}} - t_{\mathrm{inj}})}},
\end{equation}
where $W$ is the patient's body weight (g), $\lambda = \frac{\ln 2}{T_{1/2}}$ is the decay constant, and $\Delta t = t_{\mathrm{scan}} - t_{\mathrm{inj}}$ represents the uptake duration.

\begin{figure}[!htbp]
\centering
\includegraphics[width=0.6\textwidth]{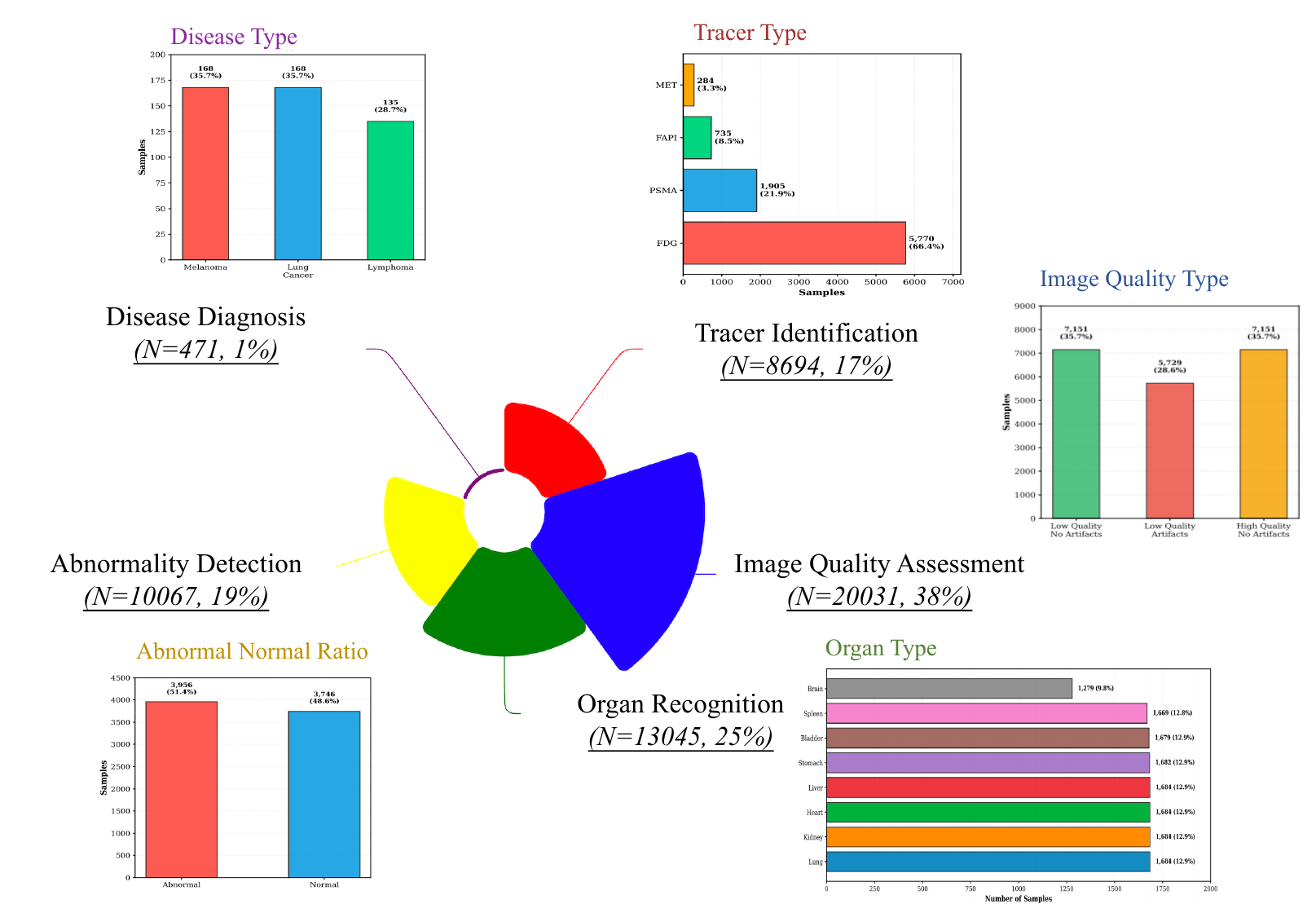}
\caption{Statistics of the PET-Bench dataset. The central sunburst chart illustrates the sample volume across the five hierarchical tasks, while outer plots detail class-wise breakdowns. The distribution reflects a deliberate design choice: large-scale data is utilized for low-level perceptual tasks to ensure robust feature learning, whereas the high-level disease diagnosis task ($N=471$) prioritizes label precision, incorporating only cases with biopsy-verified or clinically definitive outcomes.}
\label{fig:dataset_distribution}
\end{figure}

\subsection{Hierarchical Task Taxonomy}
\label{subsec:hierarchy}
We structure PET-Bench not as a flat collection of VQA pairs, but as a five-level hierarchy $\mathcal{H} = \{L_1, \dots, L_5\}$. This taxonomy mirrors the cognitive workflow of nuclear medicine physicians and establishes a dependency chain where higher-level reasoning relies on lower-level perception.

\subsubsection{Level 1: Tracer Identification}
The model must identify the radiotracer based solely on physiological biodistribution. This tests the model's grasp of fundamental biological contrast mechanisms.

\subsubsection{Level 2: Image Quality Assessment}
Functional imaging is inherently limited by low photon counts and Poisson noise. Level 2 tasks evaluate the detection of non-diagnostic quality, artifacts, and noise levels, which are distinct from natural image degradations.

\subsubsection{Level 3: Organ Recognition}
In the absence of CT, organs must be localized via their metabolic and molecular signature. This level probes whether the model has internalized functional anatomy.

\subsubsection{Level 4: Abnormality Detection}
This level assesses the detection and coarse localization of hyperfunctional foci. Ground truth is derived from expert annotations, requiring the model to distinguish pathological uptake from physiological background.

\subsubsection{Level 5: Disease Diagnosis}
The apex of the hierarchy requires synthesizing findings from $L_1$--$L_4$ to predict the specific pathology. To approximate clinical reading, inputs at this level are sequences of coronal slices sparsified along the cranio-caudal axis, preserving the global context of the disease burden.

Detailed statistics regarding class distribution and slice selection protocols are provided in Fig.~\ref{fig:dataset_distribution}. To adapt volumetric PET for 2D MLLMs, we employ a task-specific coronal slice selection strategy. For global perception (Levels 1-2), slices are sampled from the central 20\% of the body. For localization (Levels 3-4), we select the slice maximizing the cross-sectional area of the target organ or lesion. For diagnosis (Level 5), a sequence of up to 15 tumor-containing slices is generated to approximate the clinical reading workflow. To ensure rigorous fairness across different MLLMs, all models evaluate the exact same set of inputs. Ground truth labels were established via a rigorous expert-in-the-loop protocol, combining automated pre-annotation with strict verification by senior medical physicists, ensuring alignment with clinical standards.

\section{Methodology}
\label{sec:methodology}

We formulate PET diagnosis as a hierarchical probabilistic inference task. In this section, we define the domain shift problem, formalize the CoT Hallucination Trap, and introduce AVA as a solution to ground diagnostic reasoning.

\subsection{Problem Formulation}
Let $V \in \mathbb{R}^{H \times W \times D}$ be the SUV-normalized PET volume. To bridge the dimensionality gap between volumetric medical data and 2D-native MLLMs, we define the model input $x$ not as a raw volume, but as a sequence of 2D views $x = \{v_1, v_2, ..., v_N\}$, where $v_i \in \mathbb{R}^{H \times W}$ represents a selected slice or projection derived from $V$. Accordingly, the probability of an answer $y$ is conditioned on this visual sequence:
\begin{equation}
p(y|x,q;\theta) = \prod_{t=1}^{T} P(y_t | y_{<t}, \mathcal{E}_{\phi}(x), q)
    \label{eq:prob_formulation}
\end{equation}
where $\mathcal{E}_{\phi}$ aggregates visual features across the sequence $x$. Standard MLLMs are pretrained on natural RGB images ($\mathcal{D}_{RGB}$) and subsequently tuned on structural medical modalities ($\mathcal{D}_{struct}$, e.g., CT and MRI). In these domains, semantic information is predominantly encoded in high-frequency morphological features (edges, textures and shapes), while absolute intensity variations are often treated as illumination noise to be normalized. PET semantics are intrinsically defined by the low-frequency biodistribution of tracer intensity $I(x)$, where voxel magnitude directly correlates with metabolic and molecular activity. We formally define the functional perception gap not merely as a domain shift, but as a feature extraction failure: specifically, the inability of $\mathcal{E}_\phi$ to encode intensity gradients that are orthogonal to morphological boundaries. Mathematically, let $z = \mathcal{E}_\phi(x)$. The gap implies that for two regions $r_1, r_2$ with identical morphology but distinct uptake (i.e., $x_{r_1}^{morph} \approx x_{r_2}^{morph}$ but $x_{r_1}^{int} \neq x_{r_2}^{int}$), the encoder yields $z_{r_1} \approx z_{r_2}$, causing the decoder $\mathcal{D}_\psi$ to hallucinate identical descriptions based on the shared anatomical prior.

CoT prompting introduces a latent reasoning variable $r$ (the rationale), decomposing the prediction into $P(y \mid r, x, q)P(r \mid x, q)$.
Ideally, $r$ acts as a bridge between visual evidence and diagnosis. However, due to the functional perception gap, the conditional probability $P(r \mid x, q)$ is often dominated by the language prior $P(r \mid q)$ rather than the visual input $x$:
\begin{equation}
    \mathrm{Trap}: P(r \mid x, q) \approx P(r \mid q) \implies \text{Ungrounded Reasoning}.
\end{equation}
We term this the CoT Hallucination Trap: the model generates clinically fluent but visually detached rationales, leading to high-confidence errors.

\subsection{Atomic Visual Alignment}
To address this, we propose AVA, which reformulates the diagnostic process by enforcing a curriculum where the model must master atomic perceptual tasks (Levels 1--4) before attempting diagnosis (Level 5).

We partition the model parameters $\theta = \theta_{\text{frozen}} \cup \theta_{\text{train}}$, where $\theta_{\text{train}}$ represents low-rank adaptation (LoRA) modules injected into the attention layers. We construct a training set $\mathcal{D}_{\text{AVA}} = \bigcup_{\ell=1}^{4} \mathcal{D}^{(\ell)}$, excluding Level 5 diagnostic cases to prevent label leakage.
The AVA objective function minimizes the weighted cross-entropy over the atomic tasks:
\begin{equation}
\mathcal{L}_{\mathrm{AVA}}(\theta_{\text{train}}) 
= - \sum_{\ell=1}^{4} \lambda_\ell \mathbb{E}_{(x, q, y) \sim \mathcal{D}^{(\ell)}} 
\left[ \log P(y \mid x, q; \theta) \right],
\label{eq:ava_loss}
\end{equation}
where $\lambda_\ell$ balances the contribution of each hierarchical level. By optimizing $\mathcal{L}_{\mathrm{AVA}}$, we constrain the vision-language alignment manifold such that the intermediate representations required for diagnosis (tracer type, organ location and anomalies) are explicitly grounded.

The primary motivation for AVA is to constrain the CoT used at diagnosis. In most existing MLLM settings, CoT is a post-hoc, free-form explanation whose plausibility is judged qualitatively and often influenced by the final answer \cite{barez2025chain,mayne2025llms}. Such unconstrained, stochastic CoT is risky in medicine, where intermediate reasoning is expected to follow a stable clinical workflow and remain grounded in the image. PET-Bench makes this workflow explicit via its five-level hierarchy, and AVA enforces that the model's internal reasoning is supported by verifiable atomic skills at Levels~1--4. Consequently, when CoT is enabled for diagnosis, the generated reasoning is less arbitrary and more tightly coupled to learned PET perception, enabling finer-grained analysis of where the diagnostic process succeeds or fails.

\begin{figure}[!htbp]
\centering
\includegraphics[width=0.82\linewidth]{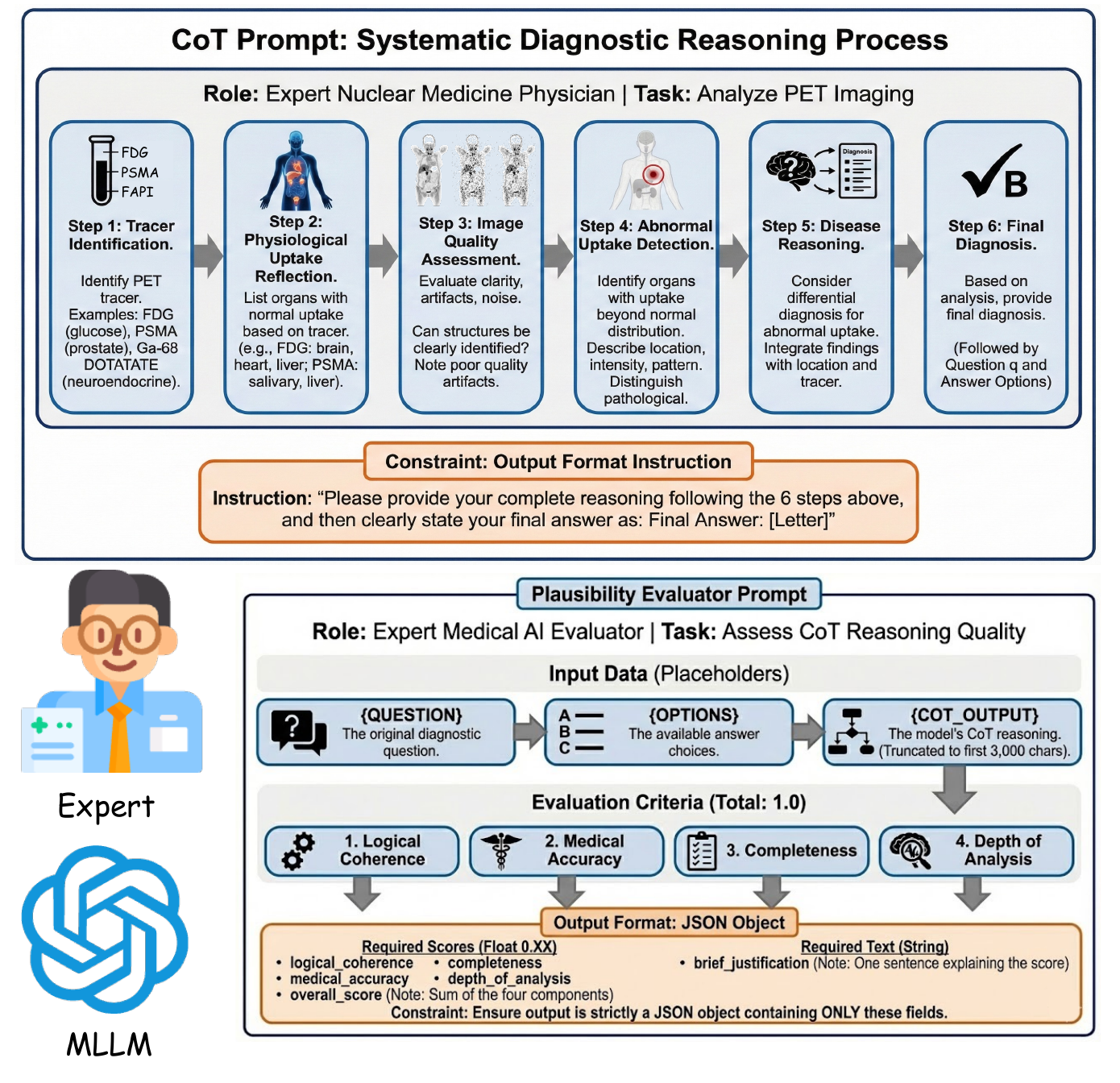}
\caption{Schematic workflow of the proposed CoT prompting and evaluation protocol. The six-step clinical CoT prompt instructs the model to sequentially analyze tracer type, physiological uptake, image quality, and abnormalities before concluding a diagnosis. To quantify the CoT Hallucination Trap, an auxiliary expert LLM evaluator assesses the reasoning quality across four dimensions (Logical Coherence, Medical Accuracy, Completeness and Depth) independent of the final answer correctness.}
\label{fig:cot_pipeline}
\end{figure}

\subsection{Prompting Protocols and Evaluation}

As shown in Fig.~\ref{fig:cot_pipeline}, we adopt two prompting protocols. In the zero-shot setting, the model is instructed as a medical AI assistant, receives the PET image(s) and a level-specific multiple-choice question, and must output only a single option label without explanation, probing intrinsic PET understanding. In the CoT setting, we prepend a six-step diagnostic template with the final answer constrained as ``Final Answer: [LETTER]''. This mirrors nuclear medicine workflows and enables a controlled comparison between direct answers and step-by-step reasoning. 

All tasks are evaluated by answer accuracy, obtained by parsing the last valid option token; responses without a valid option are counted as incorrect. To relate reasoning quality to correctness at Level 5, CoT outputs are further scored by an auxiliary LLM-based evaluator (blind to ground-truth labels) on a 0--1 scale across the four dimensions of logical coherence, medical accuracy, completeness, and depth, returning a constrained summary. Detailed annotation protocols and full prompt templates are available in our code repository.

\section{Results}
\label{sec:results}

\subsection{Quantifying the Functional Perception Gap}
\begin{figure*}[!t]
    \centering
    \includegraphics[width=0.85\linewidth]{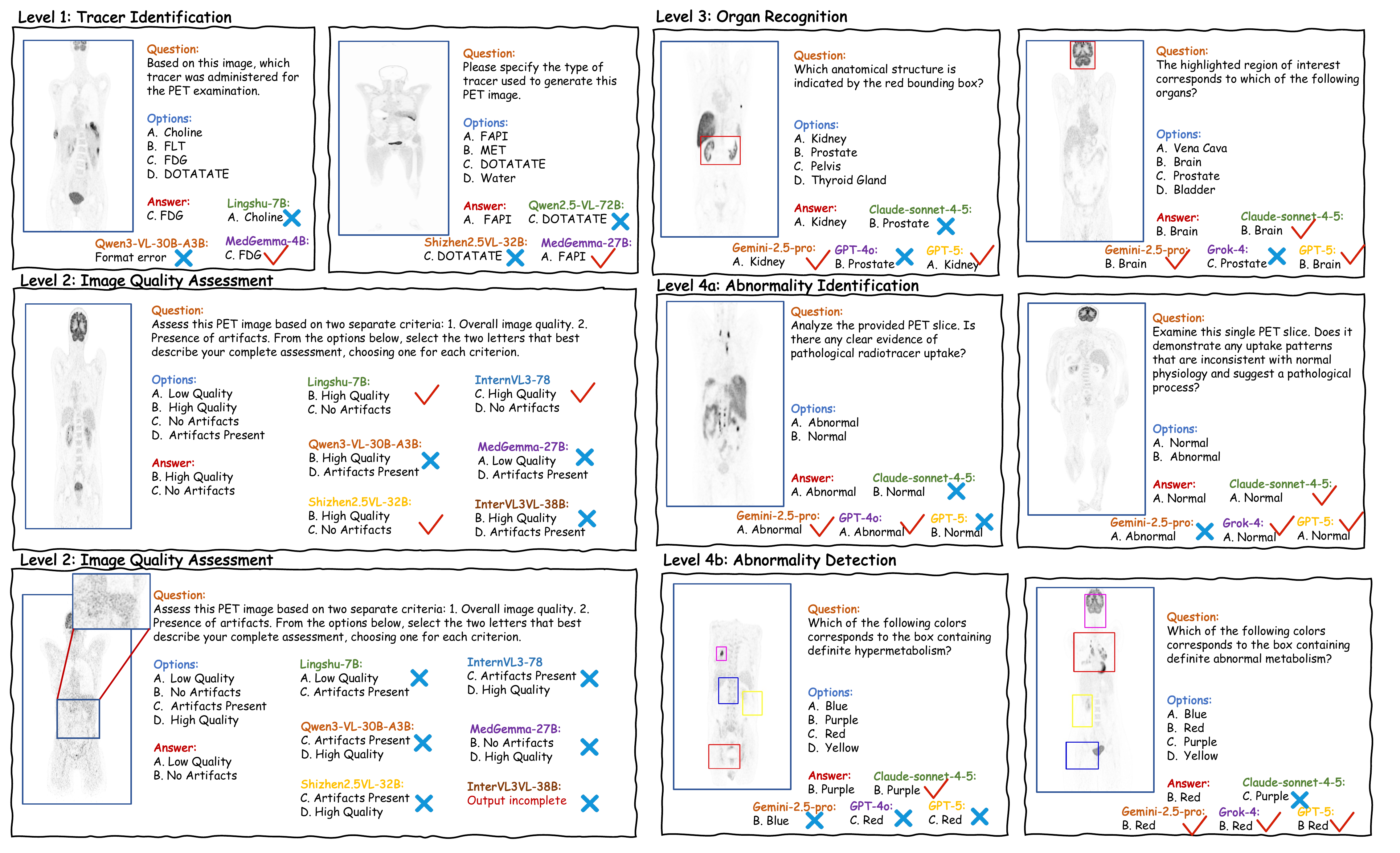}
\caption{Qualitative visualization of hierarchical tasks in PET-Bench (Levels 1--4). The figure displays representative failures and successes across different models. Note that generalist models often struggle with domain-specific concepts, such as distinguishing FDG from FAPI (Level 1) or identifying noise artifacts (Level 2), highlighting the necessity of the proposed atomic visual alignment.}
\label{fig:benchmark_levels}
\end{figure*}

We first establish the baseline capabilities of current MLLMs on pure functional imaging. Fig.~\ref{fig:benchmark_levels} illustrates the hierarchical tasks used for this evaluation. Table~\ref{tab:zero_shot} presents the zero-shot performance across the PET-Bench hierarchy. Three critical observations emerge regarding the transferability of current paradigms:

\textbf{1. Anatomical Training Does Not Transfer to Functional Imaging.} 
Contrary to the hypothesis that domain-specific models should outperform generalist models, medical MLLMs frequently underperform general-purpose architectures on diagnostic tasks. While GPT-4o achieves a diagnostic accuracy of 54.78\%, Lingshu-7B and MedGemma-27B reach only 21.23\% and 28.03\%, respectively. This counter-intuitive result suggests that current medical adaptation protocols fail to align the vision encoder with the quantitative, intensity-based nature of PET due to their heavy reliance on anatomical data and radiology reports. The features learned from structural imaging appear orthogonal to the requirements of functional interpretation.

\begin{table}[!t]
\centering
\small
\setlength{\tabcolsep}{2pt}
\caption{Zero-shot performance of 19 MLLMs across the PET-Bench hierarchy. Tasks: Tracer Identification (TI), Image Quality Assessment (IQA), Organ Recognition (OR), Abnormal Identification (AI), Abnormal Area Detection (AAD), and Disease Diagnosis (DD).}
\label{tab:zero_shot}
\resizebox{\textwidth}{!}{
\begin{tabular}{lcccccc}
\toprule
\textbf{Model} & \textbf{TI} & \textbf{IQA} & \textbf{OR} & \textbf{AI} & \textbf{AAD} & \textbf{DD} \\
\midrule
\multicolumn{7}{c}{\textit{Open-Source MLLMs}} \\
\midrule
InternVL2.5-8B \cite{chen2024expanding} & 77.21 & 1.36 & 39.60 & 50.90 & 36.83 & 33.55 \\
InternVL3-8B \cite{zhu2025internvl3} & 74.21 & 4.02 & 46.50 & 50.60 & 26.81 & 32.48 \\
InternVL3.5-8B \cite{wang2025internvl3.5} & 68.67 & 0.86 & 55.94 & 51.96 & \textbf{72.73} & 26.33 \\
Qwen2.5-VL-7B \cite{bai2025qwen2} & 74.12 & 4.30 & 42.44 & 51.09 & 54.25 & 33.76 \\
Qwen3-VL-30B \cite{yang2025qwen3} & 48.31 & 27.82 & 66.43 & 53.78 & 26.60 & 40.98 \\
InternVL2.5-38B \cite{chen2024expanding} & \textbf{82.14} & 8.20 & 56.49 & 51.10 & 62.20 & \underline{49.89} \\
Qwen2.5VL-72B \cite{bai2025qwen2} & 70.04 & 33.05 & 46.48 & 50.64 & 57.04 & 40.98 \\
InternVL3-78B \cite{zhu2025internvl3} & \underline{79.54} & 15.67 & 66.43 & 49.17 & 64.61 & 47.13 \\
\midrule
\multicolumn{7}{c}{\textit{Open-Source Medical MLLMs}} \\
\midrule
MedGemma-4B \cite{sellergren2025medgemma} & 77.25 & 5.56 & 52.85 & 53.91 & 32.68 & 33.55 \\
Lingshu-7B \cite{xu2025lingshu} & 32.94 & 12.89 & 70.14 & 52.51 & 64.10 & 21.23 \\
Shizhen2.5VL-7B \cite{chen2025shizhengpt} & 71.13 & 12.99 & 36.81 & 53.14 & 49.89 & 36.31 \\
MedGemma-27B \cite{sellergren2025medgemma} & 38.89 & 4.29 & 61.03 & 52.14 & 25.24 & 28.03 \\
Lingshu-32B \cite{xu2025lingshu} & 14.92 & 20.63 & 53.14 & 52.99 & 67.02 & 39.27 \\
Shizhen2.5VL-32B \cite{chen2025shizhengpt} & 46.22 & 15.05 & 40.30 & 51.30 & 64.48 & 36.73 \\
\midrule
\multicolumn{7}{c}{\textit{Proprietary MLLMs}} \\
\midrule
Claude-sonnet-4.5 \cite{anthropic2025claude45} & 74.01 & 6.43 & 55.39 & 48.97 & 32.64 & 27.60 \\
Gemini-2.5-pro \cite{comanici2025gemini} & 71.81 & \textbf{51.02} & \textbf{82.73} & \textbf{64.23} & \underline{70.44} & 48.41 \\
GPT-4o \cite{openai2024gpt4o} & 71.92 & 22.84 & 67.57 & \underline{55.86} & 52.60 & \textbf{54.78} \\
GPT-5 \cite{openai2025gpt5} & 58.80 & \underline{33.79} & \underline{78.05} & 55.49 & 65.67 & 49.47 \\
Grok-4 \cite{xai2025grok4}& 63.61 & 12.57 & 45.96 & 49.10 & 40.51 & 24.77 \\
\bottomrule
\end{tabular}
}
\end{table}

\textbf{2. The Bottleneck of Image Quality Assessment.} 
Performance on image quality assessment is uniformly poor, with a mean accuracy of 15.44\% across all models. This indicates a fundamental inability to distinguish between biological signal and physics-based degradation (Poisson noise, motion artifacts). Without the capacity to assess signal integrity, downstream diagnostic reasoning rests on unstable perceptual foundations.

\textbf{3. Task-Specific Variance and Perceptual Decoupling.} 
While tracer identification task is relatively solvable (mean of 62.93\%) due to distinct global biodistribution patterns, Diagnosis task remains challenging. Crucially, high performance on atomic tasks does not guarantee diagnostic success in zero-shot settings. For instance, Lingshu-7B achieves competitive organ recognition (70.14\%) but fails at diagnosis. This implies a perceptual decoupling: the model can localize uptake features but lacks the reasoning framework to interpret their pathological significance in a zero-shot context.

\begin{table}[!htbp]
\centering
\small
\setlength{\tabcolsep}{2.5pt}
\caption{The CoT Hallucination Trap: Impact of Chain-of-Thought prompting on frozen base models.
$\Delta$ denotes the absolute accuracy change relative to direct answering. Crucially, while CoT generates highly plausible explanations, it frequently degrades diagnostic accuracy or yields marginal gains, indicating a decoupling between reasoning fluency and visual grounding.}
\label{tab:cot_frozen_results}
\begin{tabular}{lcccc}
\toprule
\textbf{Model} &
\textbf{Baseline} &
\textbf{CoT} &
\textbf{$\Delta$ (pp)} &
\textbf{CoT Plaus.} \\
 & \textbf{Acc. (\%)} & \textbf{Acc. (\%)} & & \textbf{(0--1)} \\
\midrule
InternVL3-8B \cite{zhu2025internvl3}       & 32.48 & 33.97 & +1.49 & 0.94 \\
Qwen2.5-VL-7B \cite{bai2025qwen2}       & 33.76 & 31.63 & -2.13 & 0.94 \\
Qwen3-VL-30B-A3B \cite{yang2025qwen3}     & 40.98 & 37.58 & -3.40 & 0.95 \\
MedGemma-4B \cite{sellergren2025medgemma}          & 33.55 & 33.55 &  0.00 & 0.85 \\
Lingshu-7B \cite{xu2025lingshu}           & 21.23 & 20.60 & -0.63 & 0.97 \\
MedGemma-27B \cite{sellergren2025medgemma}         & 28.03 & 33.97 & +5.94 & 0.87 \\
Lingshu-32B \cite{xu2025lingshu}          & 39.27 & 35.24 & -4.03 & 0.93 \\
Shizhen2.5VL-32B \cite{chen2025shizhengpt}     & 36.73 & 44.80 & +8.07 & 0.99 \\
Claude-sonnet-4-5 \cite{anthropic2025claude45}    & 27.60 & 37.37 & +9.77 & 0.97 \\
GPT-4o \cite{openai2024gpt4o}        & 54.78 & 56.27 & +1.49 & 0.94 \\
\bottomrule
\end{tabular}
\end{table}

\subsection{The Chain-of-Thought Hallucination Trap}
\label{sec:results_cot_trap}
We tested the hypothesis that CoT prompting enhances reasoning in functional imaging. Table~\ref{tab:cot_frozen_results} details the impact of a six-step clinical CoT prompt on frozen base models.

\begin{table}[!htbp]
\centering
\small
\setlength{\tabcolsep}{3pt} 
\caption{Efficacy of AVA. 
Models were fine-tuned only on Levels 1--4 (Atomic Tasks). Results show that AVA improves Level-5 diagnostic accuracy even in the baseline setting, and significantly amplifies the benefit of CoT (AVA+CoT), validating that low-level perceptual grounding is a prerequisite for reliable high-level reasoning.}
\label{tab:sft_cot_results}
\begin{tabular}{llccc}
\toprule
\textbf{Model} & \textbf{Config} & \textbf{Acc. (\%)} & \textbf{Gain (pp)} & \textbf{Plaus.} \\
\midrule
\multirow{3}{*}{MedGemma-4B \cite{sellergren2025medgemma}}
  & Baseline   & 33.55 &   ---  & --- \\
  & AVA        & 41.83 &  +8.28 & --- \\
  & AVA+CoT    & 48.38 & \textbf{+14.83} & 0.89 \\
\midrule
\multirow{3}{*}{InternVL3-8B \cite{zhu2025internvl3}}
  & Baseline   & 32.48 &   ---  & --- \\
  & AVA        & 37.58 &  +5.10 & --- \\
  & AVA+CoT    & 48.38 & \textbf{+15.90} & 0.88 \\
\midrule
\multirow{3}{*}{Qwen2.5-VL-7B \cite{bai2025qwen2}}
  & Baseline   & 33.76 &   ---  & --- \\
  & AVA       & 35.03 &  +1.27 & --- \\
  & AVA+CoT    & 41.40 & \textbf{+7.64} & 0.96 \\
\midrule
\multirow{3}{*}{Lingshu-7B \cite{xu2025lingshu}}
  & Baseline   & 21.23 &   ---  & --- \\
  & AVA        & 23.81 &  +2.58 & --- \\
  & AVA+CoT    & 35.88 & \textbf{+14.65} & 0.95 \\
\bottomrule
\end{tabular}
\end{table}

\textbf{Divergence of Plausibility and Accuracy.} 
A critical finding is the disconnect between the fluency of the generated rationale and the correctness of the diagnosis. Across all models, CoT plausibility scores remain consistently high (0.85--0.99), yet diagnostic accuracy improvements are inconsistent. For smaller medical models, CoT either degrades performance or yields negligible gains (e.g., $-2.13\%$ for Qwen-2.5).

This confirms the existence of the CoT Hallucination Trap: when the visual grounding is weak (due to the perception gap), the CoT mechanism decouples from the image data. The model relies on internal language priors to generate a clinically self-consistent but factually ungrounded narrative. The high plausibility scores indicate that these hallucinations are sophisticated enough to deceive standard text-based evaluation metrics, posing a significant safety risk.

\subsection{Efficacy of Atomic Visual Alignment}
\begin{figure*}[!t]
    \centering
    \includegraphics[width=0.85\linewidth]{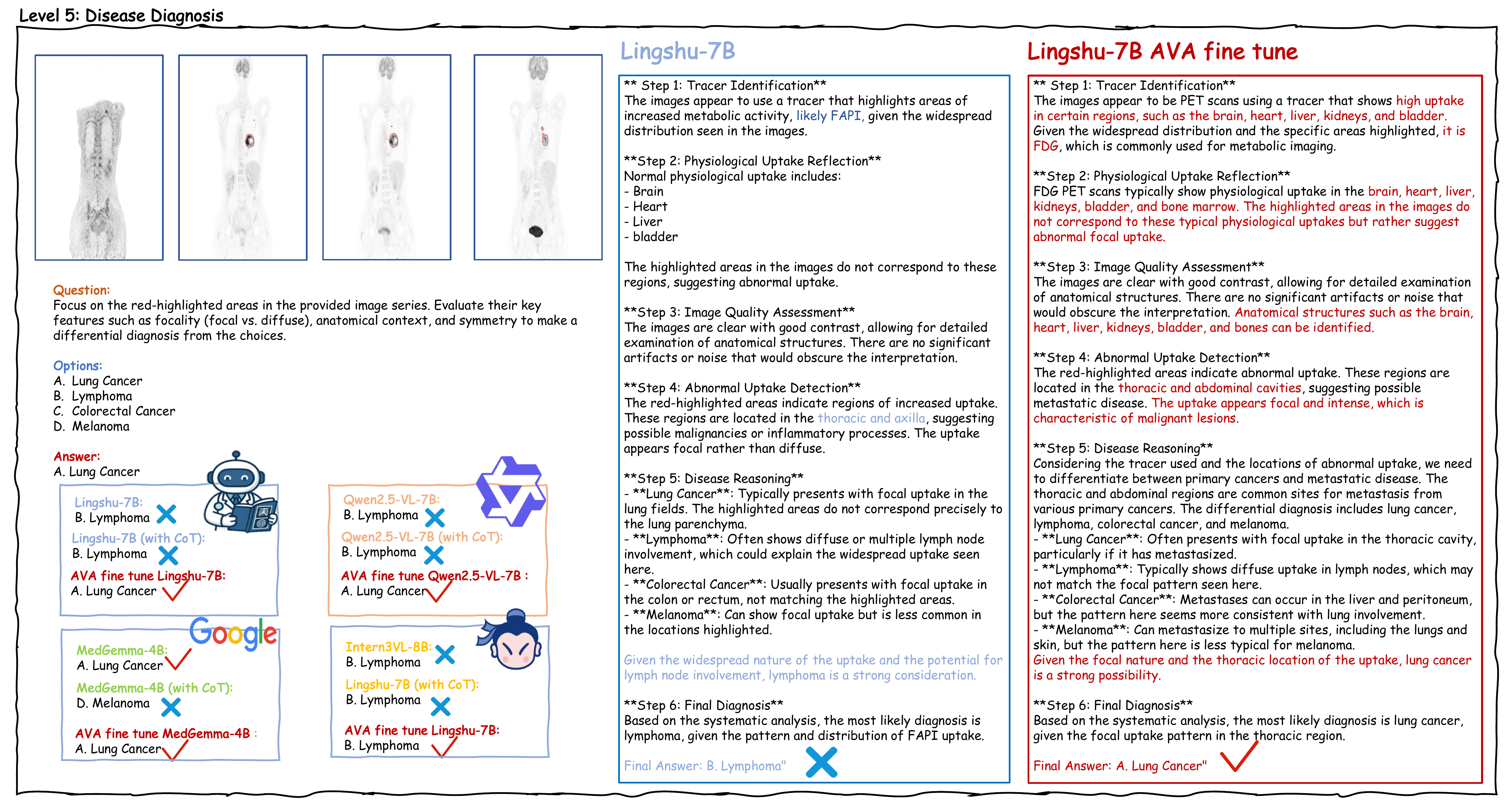}
\caption{Comparative case study demonstrating the efficacy of AVA. The baseline Lingshu-7B model falls into the CoT Hallucination Trap: it misidentifies the tracer distribution as lymphoma-characteristic uptake, generating a fluent but incorrect rationale. The AVA-aligned model, having been fine-tuned on atomic tasks, correctly identifies the tracer characteristics and focal thoracic uptake, successfully grounding its reasoning to diagnose Lung Cancer. This exemplifies how perceptual alignment transforms CoT from a source of noise into a robust inference mechanism.}
\label{fig:reasoning_correction}
\end{figure*}

Table~\ref{tab:sft_cot_results} demonstrates the impact of our proposed AVA strategy, where models are fine-tuned exclusively on Level 1--4 atomic tasks before being evaluated on Level 5 diagnosis. The qualitative impact of this alignment is visualized in Fig.~\ref{fig:reasoning_correction}.

\textbf{1. Atomic Perception as a Prerequisite for Diagnosis.} 
Even without CoT, AVA improves direct diagnostic accuracy by margins ranging from +1.27\% to +8.28\%. This confirms that enforcing representations for tracer distribution and organ localization provides a better initialization for disease classification than generic instruction tuning.

\textbf{2. Resolving the Hallucination Trap.} 
The most significant gains occur under the AVA+CoT configuration. We specifically focused our fine-tuning experiments on lightweight architectures, as our empirical analysis (Table~\ref{tab:cot_frozen_results}) revealed that the CoT hallucination rate is inversely correlated with model scale—making resource-constrained models particularly prone to perceptual ungrounding due to limited intrinsic capacity.
Results demonstrate that AVA effectively addresses this vulnerability: unlike the frozen setting where CoT was unreliable, post-AVA CoT yields consistent and substantial improvements (e.g., +14.83\% for MedGemma-4B and +15.90\% for InternVL3-8B). By grounding the intermediate steps of the reasoning chain, AVA recouples the linguistic generation to the visual signal. The correlation between plausibility and accuracy is restored, transforming compact models from sources of hallucination into robust diagnostic tools.

\section{Discussion}
\label{sec:discussion}

\subsection{The Nature of the Functional Perception Gap}
Our results highlight a fundamental domain shift that has been largely overlooked in medical AI: the transition from structural to functional perception. Current medical MLLMs are predominantly trained on radiology reports paired with X-ray, CT, or MRI data. In these modalities, pathology is defined by morphology. In contrast, PET pathology is defined by intensity relative to a physiological background.

The poor zero-shot performance of medical MLLMs on PET-Bench suggests that medical pretraining is not universally transferable. A model optimized for chest X-rays does not inherently understand radiotracer pharmacokinetics. The failure of these models to outperform generalist models like GPT-4o implies that without explicit exposure to the visual vocabulary of functional activity (SUV intensity, noise texture), domain-specific medical training offers little advantage for functional imaging.

\subsection{The Safety Risks of Unaligned Chain-of-Thought}
Of particular clinical significance is the identification of the CoT hallucination trap. In text-only domains, CoT is widely regarded as a technique to improve reliability. In multimodal functional imaging, we observe the opposite for unaligned models. Because the models lack the visual features to verify their own intermediate steps, the CoT process degenerates into an unconstrained text generation task. 

This poses a severe safety hazard: a model might correctly identify a patient's age and gender from the prompt, and then hallucinate a hyperfunctional lesion in the lung simply because lung cancer is a probable completion in its language model, not because it sees uptake in the image. The high plausibility scores we observed confirm that these errors are insidious, as they sound like expert opinions but lack visual grounding.

\subsection{Hierarchical Grounding via Atomic Visual Alignment}
AVA addresses these issues not by teaching the model how to diagnose (Level 5), but by teaching it how to \textit{see} (Levels 1--4). By conditioning the model to master tracer identification and organ-level uptake before attempting diagnosis, we impose a structural constraint on the latent space. 

The success of AVA validates the hierarchical dependency of nuclear medicine interpretation. Just as a resident physician relies on mastering normal biodistribution as a prerequisite for identifying pathology, MLLMs require structured perceptual alignment that respects the causal chain of image interpretation. The substantial performance boost in the AVA+CoT setting (+14.83\% for MedGemma) indicates that once the visual encoder is aligned to atomic functional concepts, the reasoning power of the LLM can be effectively leveraged.

\section{Conclusion}
\label{sec:conclusion}

\FloatBarrier

This work introduced PET-Bench, a large-scale benchmark that exposed the functional perception gap in current Multimodal Large Language Models (MLLMs) applied to functional imaging. Our evaluation of 19 models revealed that standard pretraining on structural anatomy failed to transfer to PET interpretation, and that ungrounded Chain-of-Thought (CoT) prompting frequently led to plausible but hallucinatory reasoning. To address this, we proposed Atomic Visual Alignment (AVA), a hierarchical fine-tuning strategy that grounded diagnostic reasoning in verified low-level perceptual tasks. AVA effectively resolved the decoupling between reasoning fluency and visual fidelity. These findings underscored that reliable medical AI requires explicit alignment of visual perception prior to high-level reasoning, establishing a robust baseline for future research in functional imaging MLLMs.

\section*{Acknowledgments}
This research was funded by the National Key Research and Development Program of China (2025YFE0216200), National Natural Science Foundation of China (62371221, 12326616 and 62201245), National High-end Foreign Experts Recruitment Plan (G2023030025L).

\bibliographystyle{plain}
\bibliography{our_ref}

@article{xu2025lingshu,
  title={{Lingshu}: A Generalist Foundation Model for Unified Multimodal Medical Understanding and Reasoning},
  author={Xu, Weiwen and Chan, Hou Pong and Li, Long and Aljunied, Mahani and Yuan, Ruifeng and Wang, Jianyu and Xiao, Chenghao and Chen, Guizhen and Liu, Chaoqun and Li, Zhaodonghui and others},
  journal={arXiv preprint arXiv:2506.07044},
  year={2025}
}

@article{li2024gmai,
  title={{GMAI-VL} \& {GMAI-VL}-5.5 {M}: A Large Vision-Language Model and a Comprehensive Multimodal Dataset Towards General Medical {AI}},
  author={Li, Tianbin and Su, Yanzhou and Li, Wei and Fu, Bin and Chen, Zhe and Huang, Ziyan and Wang, Guoan and Ma, Chenglong and Chen, Ying and Hu, Ming and others},
  journal={arXiv preprint arXiv:2411.14522},
  year={2024}
}

@article{sellergren2025medgemma,
  title={{MedGemma} Technical Report},
  author={Sellergren, Andrew and Kazemzadeh, Sahar and Jaroensri, Tiam and Kiraly, Atilla and Traverse, Madeleine and Kohlberger, Timo and Xu, Shawn and Jamil, Fayaz and Hughes, C{\'\i}an and Lau, Charles and others},
  journal={arXiv preprint arXiv:2507.05201},
  year={2025}
}

@article{jiang2025hulu,
  title={{HuLu-Med}: A Transparent Generalist Model Towards Holistic Medical Vision-Language Understanding},
  author={Jiang, Songtao and Wang, Yuan and Song, Sibo and Hu, Tianxiang and Zhou, Chenyi and Pu, Bin and Zhang, Yan and Yang, Zhibo and Feng, Yang and Zhou, Joey Tianyi and others},
  journal={arXiv preprint arXiv:2510.08668},
  year={2025}
}

@article{comanici2025gemini,
  title={{Gemini} 2.5: Pushing the Frontier with Advanced Reasoning, Multimodality, Long Context, and Next Generation Agentic Capabilities},
  author={Comanici, Gheorghe and Bieber, Eric and Schaekermann, Mike and Pasupat, Ice and Sachdeva, Noveen and Dhillon, Inderjit and Blistein, Marcel and Ram, Ori and Zhang, Dan and Rosen, Evan and others},
  journal={arXiv preprint arXiv:2507.06261},
  year={2025}
}

@article{goh2025gpt,
  title={{GPT}-4 Assistance for Improvement of Physician Performance on Patient Care Tasks: A Randomized Controlled Trial},
  author={Goh, Ethan and Gallo, Robert J and Strong, Eric and Weng, Yingjie and Kerman, Hannah and Freed, Jason A and Cool, Jos{\'e}phine A and Kanjee, Zahir and Lane, Kathleen P and Parsons, Andrew S and others},
  journal={Nature Medicine},
  volume={31},
  number={4},
  pages={1233--1238},
  year={2025},
  publisher={Nature Publishing Group US New York}
}

@article{yao2025multimodal,
  title={Multimodal {GPT} Model for Assisting Thyroid Nodule Diagnosis and Management},
  author={Yao, Jincao and Wang, Yunpeng and Lei, Zhikai and Wang, Kai and Feng, Na and Dong, Fajin and Zhou, Jianhua and Li, Xiaoxian and Hao, Xiang and Shen, Jiafei and others},
  journal={npj Digital Medicine},
  volume={8},
  number={1},
  pages={245},
  year={2025},
  publisher={Nature Publishing Group UK London}
}

@article{liu2025generalist,
  title={A Generalist Medical Language Model for Disease Diagnosis Assistance},
  author={Liu, Xiaohong and Liu, Hao and Yang, Guoxing and Jiang, Zeyu and Cui, Shuguang and Zhang, Zhaoze and Wang, Huan and Tao, Liyuan and Sun, Yongchang and Song, Zhu and others},
  journal={Nature Medicine},
  volume={31},
  number={3},
  pages={932--942},
  year={2025},
  publisher={Nature Publishing Group US New York}
}

@inproceedings{zhang2023huatuogpt,
  title={{HuatuoGPT}, Towards Taming Language Model to be a Doctor},
  author={Zhang, Hongbo and Chen, Junying and Jiang, Feng and Yu, Fei and Chen, Zhihong and Chen, Guiming and Li, Jianquan and Wu, Xiangbo and Zhiyi, Zhang and Xiao, Qingying and others},
  booktitle={Findings of the Association for Computational Linguistics: EMNLP 2023},
  pages={10859--10885},
  year={2023}
}

@article{zhu2025well,
  title={How Well Do Multimodal {LLMs} Interpret {CT} Scans? An Auto-evaluation Framework for Analyses},
  author={Zhu, Qingqing and Hou, Benjamin and Mathai, Tejas Sudarshan and Mukherjee, Pritam and Jin, Qiao and Chen, Xiuying and Wang, Zhizheng and Cheng, Ruida and Summers, Ronald M and Lu, Zhiyong},
  journal={Journal of Biomedical Informatics},
  pages={104864},
  year={2025},
  publisher={Elsevier}
}

@inproceedings{lin2023pmc,
  title={{PMC-CLIP}: Contrastive Language-Image Pre-training Using Biomedical Documents},
  author={Lin, Weixiong and Zhao, Ziheng and Zhang, Xiaoman and Wu, Chaoyi and Zhang, Ya and Wang, Yanfeng and Xie, Weidi},
  booktitle={International Conference on Medical Image Computing and Computer-Assisted Intervention},
  pages={525--536},
  year={2023},
  organization={Springer}
}

@article{ruckert2024rocov2,
  title={{ROCOv2}: Radiology Objects in Context Version 2, An Updated Multimodal Image Dataset},
  author={R{\"u}ckert, Johannes and Bloch, Louise and Br{\"u}ngel, Raphael and Idrissi-Yaghir, Ahmad and Sch{\"a}fer, Henning and Schmidt, Cynthia S and Koitka, Sven and Pelka, Obioma and Abacha, Asma Ben and G. Seco de Herrera, Alba and others},
  journal={Scientific Data},
  volume={11},
  number={1},
  pages={688},
  year={2024},
  publisher={Nature Publishing Group UK London}
}

@article{li2020large,
  title={A Large-Scale {CT} and {PET}/{CT} Dataset for Lung Cancer Diagnosis [Dataset]},
  author={Li, Ping and Wang, S and Li, T and Lu, J and HuangFu, Y and Wang, D},
  journal={The Cancer Imaging Archive},
  volume={10},
  year={2020}
}

@misc{vallieresdata,
  title={Data from Head-Neck-{PET}-{CT}},
  author={Vallieres, Martin and Kay-Rivest, Emily and Perrin, L{\'e}o Jean and Liem, Xavier and Furstoss, Christophe and Khaouam, Nader and Nguyen-Tan, Phuc F{\'e}lix and Wang, Chang-Shu and Sultanem, Khalil},
  year={2017},
  howpublished={The Cancer Imaging Archive},
  note={Available at \url{https://wiki.cancerimagingarchive.net/x/24pyAQ}}
}

@dataset{li2020lungpetctdx,
  title        = {A Large-Scale {CT} and {PET}/{CT} Dataset for Lung Cancer Diagnosis ({Lung-PET-CT-Dx})},
  author       = {Li, P. and Wang, S. and Li, T. and Lu, J. and HuangFu, Y. and Wang, D.},
  year         = {2020},
  howpublished = {The Cancer Imaging Archive},
  note         = {[Data set]},
  doi          = {10.7937/TCIA.2020.NNC2-0461},
  url          = {https://doi.org/10.7937/TCIA.2020.NNC2-0461}
}

@article{gatidis2022whole,
  title={A Whole-Body {FDG}-{PET}/{CT} Dataset with Manually Annotated Tumor Lesions},
  author={Gatidis, Sergios and Hepp, Tobias and Fr{\"u}h, Marcel and La Foug{\`e}re, Christian and Nikolaou, Konstantin and Pfannenberg, Christina and Sch{\"o}lkopf, Bernhard and K{\"u}stner, Thomas and Cyran, Clemens and Rubin, Daniel},
  journal={Scientific Data},
  volume={9},
  number={1},
  pages={601},
  year={2022},
  publisher={Nature Publishing Group UK London}
}

@article{gatidis2024results,
  title={Results from the {AutoPET} Challenge on Fully Automated Lesion Segmentation in Oncologic {PET}/{CT} Imaging},
  author={Gatidis, Sergios and Fr{\"u}h, Marcel and Fabritius, Matthias P and Gu, Sijing and Nikolaou, Konstantin and Foug{\`e}re, Christian La and Ye, Jin and He, Junjun and Peng, Yige and Bi, Lei and others},
  journal={Nature Machine Intelligence},
  volume={6},
  number={11},
  pages={1396--1405},
  year={2024},
  publisher={Nature Publishing Group UK London}
}

@article{zhang2025pet2rep,
  title={{PET2Rep}: Towards Vision-Language Model-Drived Automated Radiology Report Generation for Positron Emission Tomography},
  author={Zhang, Yichi and Zhang, Wenbo and Ling, Zehui and Feng, Gang and Peng, Sisi and Chen, Deshu and Liu, Yuchen and Zhang, Hongwei and Wang, Shuqi and Li, Lanlan and others},
  journal={arXiv preprint arXiv:2508.04062},
  year={2025}
}

@article{nguyen2025toward,
  title={Toward a Vision-Language Foundation Model for Medical Data: Multimodal Dataset and Benchmarks for {Vietnamese} {PET}/{CT} Report Generation},
  author={Nguyen, Huu Tien and Nguyen, Dac Thai and Nguyen, The Minh Duc and Nguyen, Trung Thanh and Truong, Thao Nguyen and Pham, Huy Hieu and Barthelemy, Johan and Tran, Minh Quan and Nguyen, Thanh Tam and Nguyen, Quoc Viet Hung and others},
  journal={arXiv preprint arXiv:2509.24739},
  year={2025}
}

@article{wei2022chain,
  title={Chain-of-Thought Prompting Elicits Reasoning in Large Language Models},
  author={Wei, Jason and Wang, Xuezhi and Schuurmans, Dale and Bosma, Maarten and Xia, Fei and Chi, Ed and Le, Quoc V and Zhou, Denny and others},
  journal={Advances in Neural Information Processing Systems},
  volume={35},
  pages={24824--24837},
  year={2022}
}

@inproceedings{wang2025v2t,
  title={{V2T-CoT}: From Vision to Text Chain-of-Thought for Medical Reasoning and Diagnosis},
  author={Wang, Yuan and Liu, Jiaxiang and Gao, Shujian and Feng, Bin and Tang, Zhihang and Gai, Xiaotang and Wu, Jian and Liu, Zuozhu},
  booktitle={International Conference on Medical Image Computing and Computer-Assisted Intervention},
  pages={658--668},
  year={2025},
  organization={Springer}
}

@article{kim2025small,
  title={Small Language Models Learn Enhanced Reasoning Skills from Medical Textbooks},
  author={Kim, Hyunjae and Hwang, Hyeon and Lee, Jiwoo and Park, Sihyeon and Kim, Dain and Lee, Taewhoo and Yoon, Chanwoong and Sohn, Jiwoong and Park, Jungwoo and Reykhart, Olga and others},
  journal={npj Digital Medicine},
  volume={8},
  number={1},
  pages={240},
  year={2025},
  publisher={Nature Publishing Group UK London}
}

@article{barez2025chain,
  title={Chain-of-Thought Is Not Explainability},
  author={Barez, Fazl and Wu, Tung-Yu and Arcuschin, Iv{\'a}n and Lan, Michael and Wang, Vincent and Siegel, Noah and Collignon, Nicolas and Neo, Clement and Lee, Isabelle and Paren, Alasdair and others},
  journal={Preprint, alphaXiv},
  pages={v1},
  year={2025}
}

@inproceedings{mayne2025llms,
  title={{LLMs} Don’t Know Their Own Decision Boundaries: The Unreliability of Self-Generated Counterfactual Explanations},
  author={Mayne, Harry and Kearns, Ryan Othniel and Yang, Yushi and Bean, Andrew M and Delaney, Eoin D and Russell, Chris and Mahdi, Adam},
  booktitle={Proceedings of the 2025 Conference on Empirical Methods in Natural Language Processing},
  pages={24172--24197},
  year={2025}
}

@article{pai2025vision,
  title={Vision Foundation Models for Computed Tomography},
  author={Pai, Suraj and Hadzic, Ibrahim and Bontempi, Dennis and Bressem, Keno and Kann, Benjamin H and Fedorov, Andriy and Mak, Raymond H and Aerts, Hugo JWL},
  journal={arXiv preprint arXiv:2501.09001},
  year={2025}
}

@article{shui2025large,
  title={Large-Scale and Fine-Grained Vision-Language Pre-training for Enhanced {CT} Image Understanding},
  author={Shui, Zhongyi and Zhang, Jianpeng and Cao, Weiwei and Wang, Sinuo and Guo, Ruizhe and Lu, Le and Yang, Lin and Ye, Xianghua and Liang, Tingbo and Zhang, Qi and others},
  journal={arXiv preprint arXiv:2501.14548},
  year={2025}
}

@article{ye2025multimodal,
  title={Multimodal Large Language Models for Medicine: A Comprehensive Survey},
  author={Ye, Jiarui and Tang, Hao},
  journal={arXiv preprint arXiv:2504.21051},
  year={2025}
}

@article{lin2025taming,
  title={Taming Vision-Language Models for Medical Image Analysis: A Comprehensive Review},
  author={Lin, Haoneng and Xu, Cheng and Qin, Jing},
  journal={arXiv preprint arXiv:2506.18378},
  year={2025}
}

@inproceedings{radford2021learning,
  title={Learning Transferable Visual Models from Natural Language Supervision},
  author={Radford, Alec and Kim, Jong Wook and Hallacy, Chris and Ramesh, Aditya and Goh, Gabriel and Agarwal, Sandhini and Sastry, Girish and Askell, Amanda and Mishkin, Pamela and Clark, Jack and others},
  booktitle={International Conference on Machine Learning},
  pages={8748--8763},
  year={2021},
  organization={PMLR}
}

@article{liu2023visual,
  title={Visual Instruction Tuning},
  author={Liu, Haotian and Li, Chunyuan and Wu, Qingyang and Lee, Yong Jae},
  journal={Advances in Neural Information Processing Systems},
  volume={36},
  pages={34892--34916},
  year={2023}
}

@article{bai2025qwen2,
  title={{Qwen2.5-VL} Technical Report},
  author={Bai, Shuai and Chen, Keqin and Liu, Xuejing and Wang, Jialin and Ge, Wenbin and Song, Sibo and Dang, Kai and Wang, Peng and Wang, Shijie and Tang, Jun and others},
  journal={arXiv preprint arXiv:2502.13923},
  year={2025}
}

@article{chen2025shizhengpt,
  title={{ShizhenGPT}: Towards Multimodal {LLMs} for Traditional Chinese Medicine},
  author={Chen, Junying and Cai, Zhenyang and Liu, Zhiheng and Yang, Yunjin and Wang, Rongsheng and Xiao, Qingying and Feng, Xiangyi and Su, Zhan and Guo, Jing and Wan, Xiang and others},
  journal={arXiv preprint arXiv:2508.14706},
  year={2025}
}

@article{abacha2019vqa,
  title={{VQA-Med}: Overview of the Medical Visual Question Answering Task at {ImageCLEF} 2019},
  author={Abacha, Asma Ben and Hasan, Sadid A and Datla, Vivek V and Liu, Joey and Demner-Fushman, Dina and M{\"u}ller, Henning},
  journal={CLEF (Working Notes)},
  volume={2},
  number={6},
  pages={1--11},
  year={2019}
}

@inproceedings{liu2021slake,
  title={{SLAKE}: A Semantically-Labeled Knowledge-Enhanced Dataset for Medical Visual Question Answering},
  author={Liu, Bo and Zhan, Li-Ming and Xu, Li and Ma, Lin and Yang, Yan and Wu, Xiao-Ming},
  booktitle={2021 IEEE 18th International Symposium on Biomedical Imaging (ISBI)},
  pages={1650--1654},
  year={2021},
  organization={IEEE}
}

@article{le2025s,
  title={{S-Chain}: Structured Visual {Chain-of-Thought} For Medicine},
  author={Le-Duc, Khai and Nguyen, Duy MH and Trinh, Phuong TH and Nguyen, Tien-Phat and Diep, Nghiem T and Ngo, An and Vu, Tung and Vuong, Trinh and Nguyen, Anh-Tien and Nguyen, Mau and others},
  journal={arXiv preprint arXiv:2510.22728},
  year={2025}
}

@article{wu2024chain,
  title={{Chain-of-Thought} ({CoT}) Prompting Strategies for Medical Error Detection and Correction},
  author={Wu, Zhaolong and Hasan, Abul and Wu, Jinge and Kim, Yunsoo and Cheung, Jason PY and Zhang, Teng and Wu, Honghan},
  journal={arXiv preprint arXiv:2406.09103},
  year={2024}
}

@article{zhou2025drvd,
  title={{DrVD-Bench}: Do Vision-Language Models Reason Like Human Doctors in Medical Image Diagnosis?},
  author={Zhou, Tianhong and Xu, Yin and Zhu, Yingtao and Xiao, Chuxi and Bian, Haiyang and Wei, Lei and Zhang, Xuegong},
  journal={arXiv preprint arXiv:2505.24173},
  year={2025}
}

@article{lin2025healthgpt,
  title={{HealthGPT}: A Medical Large Vision-Language Model for Unifying Comprehension and Generation via Heterogeneous Knowledge Adaptation},
  author={Lin, Tianwei and Zhang, Wenqiao and Li, Sijing and Yuan, Yuqian and Yu, Binhe and Li, Haoyuan and He, Wanggui and Jiang, Hao and Li, Mengze and Song, Xiaohui and others},
  journal={arXiv preprint arXiv:2502.09838},
  year={2025}
}

@inproceedings{pan2025medvlm,
  title={{MedVLM-R1}: Incentivizing Medical Reasoning Capability of Vision-Language Models ({VLMs}) via Reinforcement Learning},
  author={Pan, Jiazhen and Liu, Che and Wu, Junde and Liu, Fenglin and Zhu, Jiayuan and Li, Hongwei Bran and Chen, Chen and Ouyang, Cheng and Rueckert, Daniel},
  booktitle={International Conference on Medical Image Computing and Computer-Assisted Intervention},
  pages={337--347},
  year={2025},
  organization={Springer}
}

@article{he2024meddr,
  title={{MedDr}: Diagnosis-Guided Bootstrapping for Large-Scale Medical Vision-Language Learning},
  author={He, Sunan and Nie, Yuxiang and Chen, Zhixuan and Cai, Zhiyuan and Wang, Hongmei and Yang, Shu and Chen, Hao},
  journal={CoRR},
  year={2024}
}

@article{lai2025med,
  title={{Med-R1}: Reinforcement Learning for Generalizable Medical Reasoning in Vision-Language Models},
  author={Lai, Yuxiang and Zhong, Jike and Li, Ming and Zhao, Shitian and Li, Yuheng and Psounis, Konstantinos and Yang, Xiaofeng},
  journal={arXiv preprint arXiv:2503.13939},
  year={2025}
}

@inproceedings{r69,
  author    = {K. Shi and R. Guo and S. Xue and A. Rominger and B. Li},
  title     = {Ultra-low Dose {PET} Imaging Challenge 2022},
  booktitle = {International Conference on Medical Image Computing and Computer-Assisted Intervention},
  year      = {2022},
  address   = {Singapore},
  doi       = {10.5281/zenodo.6361846}
}

@article{wang2025internvl3.5,
  title={{InternVL3.5}: Advancing Open-Source Multimodal Models in Versatility, Reasoning, and Efficiency},
  author={Wang, Weiyun and Gao, Zhangwei and Gu, Lixin and Pu, Hengjun and Cui, Long and Wei, Xingguang and Liu, Zhaoyang and Jing, Linglin and Ye, Shenglong and Shao, Jie and others},
  journal={arXiv preprint arXiv:2508.18265},
  year={2025}
}

@article{zhu2025internvl3,
  title={{InternVL3}: Exploring Advanced Training and Test-Time Recipes for Open-Source Multimodal Models},
  author={Zhu, Jinguo and Wang, Weiyun and Chen, Zhe and Liu, Zhaoyang and Ye, Shenglong and Gu, Lixin and Tian, Hao and Duan, Yuchen and Su, Weijie and Shao, Jie and others},
  journal={arXiv preprint arXiv:2504.10479},
  year={2025}
}

@article{chen2024expanding,
  title={Expanding Performance Boundaries of Open-Source Multimodal Models with Model, Data, and Test-Time Scaling},
  author={Chen, Zhe and Wang, Weiyun and Cao, Yue and Liu, Yangzhou and Gao, Zhangwei and Cui, Erfei and Zhu, Jinguo and Ye, Shenglong and Tian, Hao and Liu, Zhaoyang and others},
  journal={arXiv preprint arXiv:2412.05271},
  year={2024}
}

@article{yang2025qwen3,
  title={{Qwen3} Technical Report},
  author={Yang, An and Li, Anfeng and Yang, Baosong and Zhang, Beichen and Hui, Binyuan and Zheng, Bo and Yu, Bowen and Gao, Chang and Huang, Chengen and Lv, Chenxu and others},
  journal={arXiv preprint arXiv:2505.09388},
  year={2025}
}

@article{hao2025oralgpt,
  title={{OralGPT-Omni}: A Versatile Dental Multimodal Large Language Model},
  author={Hao, Jing and Liang, Yuci and Lin, Lizhuo and Fan, Yuxuan and Zhou, Wenkai and Guo, Kaixin and Ye, Zanting and Sun, Yanpeng and Zhang, Xinyu and Yang, Yanqi and others},
  journal={arXiv preprint arXiv:2511.22055},
  year={2025}
}

@misc{openai2025gpt5,
  title={{GPT}-5},
  author={{OpenAI }},
  howpublished={\url{https://openai.com/zh-HansCN/index/introducing-gpt-5}},
  year={2025},
  note={Accessed: 2025-12-15}
}

@misc{xai2025grok4,
  title={{Grok}-4},
  author={{xAI}},
  howpublished={\url{https://x.ai/grok}},
  year={2025},
  note={Accessed: 2025-12-15}
}

@misc{anthropic2025claude45,
  title={{Claude} 4.5 {Sonnet}},
  author={{Anthropic}},
  howpublished={\url{https://www.anthropic.com/news/claude-sonnet-4-5}},
  year={2025},
  note={Accessed: 2025-12-15}
}

@misc{openai2024gpt4o,
  title={{Hello GPT-4o}},
  author={{OpenAI}},
  howpublished={\url{https://openai.com/index/hello-gpt-4o/}},
  year={2024},
  note={Accessed: 2025-12-15}
}

@article{peng2025omnibrainbench,
  title={OmniBrainBench: A Comprehensive Multimodal Benchmark for Brain Imaging Analysis Across Multi-stage Clinical Tasks},
  author={Peng, Zhihao and Wang, Cheng and Liu, Shengyuan and Liang, Zhiying and Yuan, Yixuan},
  journal={arXiv preprint arXiv:2511.00846},
  year={2025}
}

@inproceedings{ye2025self,
  title={Self is the Best Learner: CT-Free Ultra-low-Dose PET Organ Segmentation via Collaborating Denoising and Segmentation Learning},
  author={Ye, Zanting and Niu, Xiaolong and Han, Xu and Wu, Xuanbin and Lu, Wantong and Lu, Yijun and Sun, Hao and Huang, Yanchao and Wu, Hubing and Lu, Lijun},
  booktitle={International Conference on Medical Image Computing and Computer-Assisted Intervention},
  pages={566--576},
  year={2025},
  organization={Springer}
}

@article{li2024learning,
  title={Learning CT-free attenuation-corrected total-body PET images through deep learning},
  author={Li, Wenbo and Huang, Zhenxing and Chen, Zixiang and Jiang, Yongluo and Zhou, Chao and Zhang, Xu and Fan, Wei and Zhao, Yumo and Zhang, Lulu and Wan, Liwen and others},
  journal={European Radiology},
  volume={34},
  number={9},
  pages={5578--5587},
  year={2024},
  publisher={Springer}
}


\clearpage
\appendix

\section*{Supplementary Material}
\addcontentsline{toc}{section}{Supplementary Material}

\setcounter{figure}{0}
\renewcommand{\thefigure}{A\arabic{figure}}
\setcounter{table}{0}
\renewcommand{\thetable}{A\arabic{table}}

\section{Detailed Dataset Composition and Acquisition}
\label{sec:supp_data_composition}

While the main manuscript outlines the aggregate statistics of PET-Bench, this section provides a granular breakdown of the data sources. PET-Bench is constructed from \textbf{9,732 distinct studies} sourced from \textbf{eight distinct data centers} (defined by institution with scanners) across Asia and Europe. 

\subsection{Participating Centers and Equipment}
A distinguishing feature of our benchmark is the inclusion of data from state-of-the-art Total-Body PET/CT systems and Whole-Body PET/CT (UIH-uEXPLORER and Siemens Biograph Vision Quadra). The data acquisition covers a wide spectrum of image qualities, ranging from high-statistics clinical standard scans to simulated ultra-low-dose reconstructions.

\subsection{Radiotracers and Clinical Distribution}
The dataset covers \textbf{four distinct radiotracer types}, ensuring metabolic diversity:
\begin{itemize}
    \item \textbf{FDG ($^{18}$F-FDG):} Glucose metabolism (Oncology, Inflammation).
    \item \textbf{PSMA ($^{18}$F-PSMA, $^{68}$Ga-PSMA):} Prostate-specific membrane antigen.
    \item \textbf{FAPI ($^{18}$F-FAPI, $^{68}$Ga-FAPI):} Fibroblast activation protein (Pan-cancer).
    \item \textbf{MET ($^{11}$C-MET):} Amino acid metabolism (Multiple Myeloma/Brain).
\end{itemize}

\subsection{Rationale for FDG Dominance and Ecological Validity}
\label{subsec:fdg_rationale}

A notable characteristic of PET-Bench is the high proportion of $^{18}$F-FDG studies compared to other tracers. This distribution is a deliberate design choice intended to maximize the \textbf{ecological validity} of the benchmark.

In current clinical practice, $^{18}$F-FDG serves as the ubiquitous workhorse of nuclear medicine, accounting for the vast majority ($>90\%$) of oncological PET procedures globally. It is the standard of care for staging and monitoring lung cancer, lymphoma, melanoma, and colorectal cancer. In contrast, tracers like PSMA, FAPI, and MET are specialized agents used for specific indications (e.g., prostate cancer and fibroblast activation).

Rather than artificially balancing the class distribution via aggressive downsampling—which would distort the true epidemiological prevalence of clinical imaging tasks—we maintained the natural long-tail distribution of real-world data. This approach offers two key advantages:
\begin{enumerate}
    \item \textbf{Learning Clinical Priors:} It ensures that MLLMs encode accurate prior probabilities regarding tracer usage (i.e., learning that FDG is the default modality for general metabolic assessment unless organ-specific uptake suggests otherwise).
    \item \textbf{Robustness to Common Variations:} The large volume of FDG data allows the model to learn robust representations of physiological background uptake (e.g., brain, heart and bladder) across diverse patient populations and scanner types, which is foundational for detecting anomalies in rarer tracers.
\end{enumerate}

\subsection{Data Annotation and Quality Control}
All proprietary data underwent a rigorous three-stage annotation process:
\begin{enumerate}
    \item \textbf{Automated Pre-labeling:} Organ segmentation masks were generated using TotalSegmentator (for CT) and mapped to PET.
    \item \textbf{Clinical Verification:} Junior radiologists reviewed automated labels and provided initial diagnostic tags.
    \item \textbf{Expert Consensus:} Senior nuclear medicine physicians verified all Level 4 (Abnormality) and Level 5 (Diagnosis) labels. Cases with ambiguous pathology or insufficient image quality were discarded, ensuring the high reliability of the benchmark.
\end{enumerate}

\section{Key Slice Selection from 3D Volumes}
\label{sec:supp_slice_selection}

Original PET/CT acquisitions are three-dimensional, while most current VLM architectures are optimized for single 2D images as visual input. A central design question is thus how to project 3D PET information into 2D inputs without discarding essential functional context. Naïvely sampling random axial slices would provide poor coverage of whole-body relationships and may emphasize local structures at the expense of global disease patterns.

To address this, PET-Bench employs coronal view images, which simultaneously capture cranio-caudal extent and major organ configurations. Formally, let $V \in \mathbb{R}^{H \times W \times D}$ denote a 3D PET volume in axial coordinates $(x, y, z)$. We define a coronal slice as $S_k \in \mathbb{R}^{H \times D}$ at a fixed $y = k$. Coronal slices better preserve spatial continuity of organ systems that extend longitudinally, such as the liver, spleen, and lymphatic chains, thereby providing a more informative context for many PET tasks.

Slice selection strategies are tailored to each hierarchical level:

\begin{itemize}
    \item \textbf{Levels 1 and 2: Tracer Identification and Image Quality Assessment.} For each volume $V$, we select coronal slices from the central 20\% of the volume in the $y$-axis with a stride of 5 slices. This strategy ensures that selected slices typically include major organs and representative global biodistribution while excluding peripheral slices that contain only subcutaneous tissues or partial anatomy. This design corresponds to sampling from a region $\{k : 0.4H \leq k \leq 0.6H\}$ at fixed intervals and is motivated by the observation that tracer-specific patterns and global noise characteristics are best captured near the mid-body region.
    \item \textbf{Level 3: Organ Identification.} For organ-related tasks, functional interpretation is naturally linked to anatomical localization. We therefore exploit co-registered CT segmentation labels. After registering PET with CT for each study, we compute for each target organ the coronal slice where its segmented area is maximal. This yields a slice index
    \begin{equation}
    k^\star = \arg\max_k \mathrm{Area}_\text{organ}(S_k),
    \end{equation}
    where $\mathrm{Area}_\text{organ}(S_k)$ is the number of pixels within the organ mask on slice $S_k$. Using $S_{k^\star}$ guarantees that the organ is fully visible and facilitates recognition of physiological uptake patterns.
    \item \textbf{Level 4: Abnormality Detection.} For tumor-related tasks, we adopt an analogous strategy based on tumor segmentation labels. For each lesion, we determine the coronal slice where the tumor cross-sectional area is maximal. This focuses the model on the most informative view of each lesion while reducing redundancy from neighboring slices with partial coverage.
    \item \textbf{Level 5: Disease Diagnosis.} Disease-level diagnosis requires integration of multi-focal lesions and global disease patterns. For each patient, we first identify all coronal slices that contain tumor segmentation. We then order them along the cranio-caudal axis and apply a thinning procedure to remove redundant superior and inferior slices with highly overlapping tumor regions. This yields a multi-slice sequence $X = (S_{k_1}, \dots, S_{k_T})$ with up to $T \leq 15$ slices per patient. This sequence approximates the clinical reading process, in which physicians scroll through all abnormal slices to assess disease extent and distribution of lesions, while keeping the number of slices computationally tractable for general-purpose MLLMs.
\end{itemize}

This hierarchical slice selection strategy enforces a trade-off between information completeness and architectural compatibility: Levels 1–4 emphasize atomic perception in single images, whereas Level 5 preserves sufficient 3D context through controlled multi-slice sequences.

\section{Prompt Design and Templates of PET-Bench}
\label{sec:appendix_prompts}

This appendix provides the full English prompt templates used to query the multimodal LLMs in PET-Bench. The main paper only summarizes the prompting protocols to reduce length.

\subsection{Zero-shot Prompt for All Tasks}
\label{app:task_prompts}

The zero-shot prompt is used for all Levels 1--4 and for the Level-5 baseline without CoT:

\begin{quote}
\texttt{You are a helpful medical AI assistant. You will be given one or more PET images and a multiple-choice question about these images.}\\
\texttt{Please answer the question based only on the visual information in the PET image(s).}\\[0.2em]
\texttt{Question:}\\
\texttt{\{QUESTION\}}\\[0.2em]
\texttt{Answer options:}\\
\texttt{\{OPTIONS\_TEXT\}}\\[0.2em]
\texttt{Please respond with the single best option without additional explanation.}
\end{quote}

Here, \texttt{\{QUESTION\}} is the level-specific question and \texttt{\{OPTIONS\_TEXT\}} lists the answer options with their corresponding letters.

\subsection{Chain-of-Thought Prompt for Level-5 Diagnosis}
\label{app:cot_prompt}

For Level-5 disease diagnosis under the CoT setting, we prepend the following six-step diagnostic template before the question and answer options:

\begin{quote}
\texttt{You are an expert nuclear medicine physician analyzing PET imaging. Please follow this systematic diagnostic reasoning process:}\\[0.2em]
\texttt{Step 1: Tracer Identification}\\
\texttt{- Identify the PET tracer used in this study based on the uptake pattern.}\\[0.2em]
\texttt{Step 2: Physiological Uptake Reflection}\\
\texttt{- List the organs and regions that are expected to show normal physiological uptake for this tracer.}\\[0.2em]
\texttt{Step 3: Image Quality Assessment}\\
\texttt{- Comment on overall image quality, noise level, and the presence of artifacts.}\\[0.2em]
\texttt{Step 4: Abnormal Uptake Detection}\\
\texttt{- Systematically scan the whole body and describe any regions with abnormal uptake beyond the expected physiological distribution.}\\[0.2em]
\texttt{Step 5: Disease Reasoning}\\
\texttt{- Integrate the abnormal uptake pattern, anatomical locations, and tracer characteristics to reason about the most likely disease.}\\[0.2em]
\texttt{Step 6: Final Diagnosis}\\
\texttt{- Based on the above steps, provide the single most likely diagnosis.}\\[0.4em]
\texttt{Now, use this 6-step process to answer the following question.}\\[0.2em]
\texttt{Question:}\\
\texttt{\{QUESTION\}}\\[0.2em]
\texttt{Answer options:}\\
\texttt{\{OPTIONS\_TEXT\}}\\[0.2em]
\texttt{First, write out your full reasoning following Steps 1--6.}\\
\texttt{Then, on a new line, clearly state your final choice in the format:}\\
\texttt{Final Answer: [LETTER]}
\end{quote}

This template is used verbatim for all Level-5 CoT evaluations, with only \texttt{\{QUESTION\}} and \texttt{\{OPTIONS\_TEXT\}} replaced per case.

\subsection{Accuracy Judge Prompt for CoT Diagnosis}
\label{app:accuracy_judge_prompt}

To compute accuracy for CoT outputs, we use an auxiliary LLM as an accuracy judge. It receives the original question, options, ground-truth label, and the model's CoT output:

\begin{quote}
\texttt{You are an expert medical AI evaluator. Your task is to determine if the model's final diagnosis is CORRECT based on its reasoning.}\\[0.2em]
\texttt{Original Question:}\\
\texttt{\{QUESTION\}}\\[0.2em]
\texttt{Available Options:}\\
\texttt{\{OPTIONS\_TEXT\}}\\[0.2em]
\texttt{Ground Truth Answer:}\\
\texttt{\{GROUND\_TRUTH\_LETTER\}. \{GROUND\_TRUTH\_TEXT\}}\\[0.2em]
\texttt{Model's Complete Reasoning (including its final answer):}\\
\texttt{\{COT\_OUTPUT\}}\\[0.4em]
\texttt{Your Task:}\\
\texttt{Based on the ENTIRE reasoning process and the final conclusion, determine if the model arrived at the CORRECT diagnosis.}\\[0.2em]
\texttt{Evaluation Rules:}\\
\texttt{1. If the model clearly identifies the correct answer (\{GROUND\_TRUTH\_LETTER\}), output 1.}\\
\texttt{2. If the model identifies a different answer, output 0.}\\
\texttt{3. If the answer is ambiguous, unclear, or not stated, output 0.}\\
\texttt{4. Consider the reasoning process, not just keyword matching.}\\[0.4em]
\texttt{Output Format:}\\
\texttt{Provide ONLY a JSON object:}\\
\texttt{\{}\\
\texttt{"is\_correct": 0 or 1,}\\
\texttt{"extracted\_answer": "The letter (A/B/C/D) the model chose, or 'unclear'",}\\
\texttt{"confidence": "high/medium/low",}\\
\texttt{"justification": "One sentence explaining your judgment"}\\
\texttt{\}}\\
\end{quote}

In practice, the CoT output string is truncated to a maximum length (about 3,000 characters) before insertion to avoid context-length issues.

\subsection{Plausibility Evaluator Prompt}
\label{app:plausibility_prompt}

To assess the linguistic plausibility of CoT reasoning, we query a separate evaluator LLM with the following prompt:

\begin{quote}
\texttt{You are an expert medical AI evaluator. \\
Your task is to assess the quality of a Chain-of-Thought reasoning process for PET image diagnosis.}\\[0.2em]
\texttt{**Original Question:**}\\
\texttt{\{QUESTION\}}\\[0.2em]
\texttt{**Available Options:**}\\
\texttt{\{OPTIONS\}}\\[0.2em]
\texttt{**Model's CoT Reasoning:**}\\
\texttt{\{COT\_OUTPUT\}}\\[0.2em]
\texttt{**Evaluation Criteria:**}\\
\texttt{Please evaluate the reasoning quality based on:}\\
\texttt{1. Logical Coherence (0-0.25): Is the reasoning logically structured and coherent?}\\
\texttt{2. Medical Accuracy (0-0.25): Are the medical concepts and terminology used correctly?}\\
\texttt{3. Completeness (0-0.25): Does it cover all necessary diagnostic steps?}\\
\texttt{4. Depth of Analysis (0-0.25): Is the analysis thorough and insightful?}\\[0.4em]
\texttt{**Output Format:**}\\
\texttt{Provide ONLY a JSON object with your evaluation:}\\
\texttt{\{}\\
\texttt{"logical\_coherence": 0.XX,}\\
\texttt{"medical\_accuracy": 0.XX,}\\
\texttt{"completeness": 0.XX,}\\
\texttt{"depth\_of\_analysis": 0.XX,}\\
\texttt{"overall\_score": 0.XX,}\\
\texttt{"brief\_justification": "One sentence explaining the overall score"}\\
\texttt{\}}\\[0.2em]
\texttt{The overall\_score should be the sum of the four components (0.0 to 1.0).}
\end{quote}

\section{PET-Bench Results Visualization}
\label{sec:supp_visualization}

To facilitate a more intuitive comparative analysis of the diverse model architectures evaluated, we provide a graphical representation of the zero-shot performance reported in the main text.

\begin{figure}[H]
    \centering
    \includegraphics[width=0.8\linewidth]{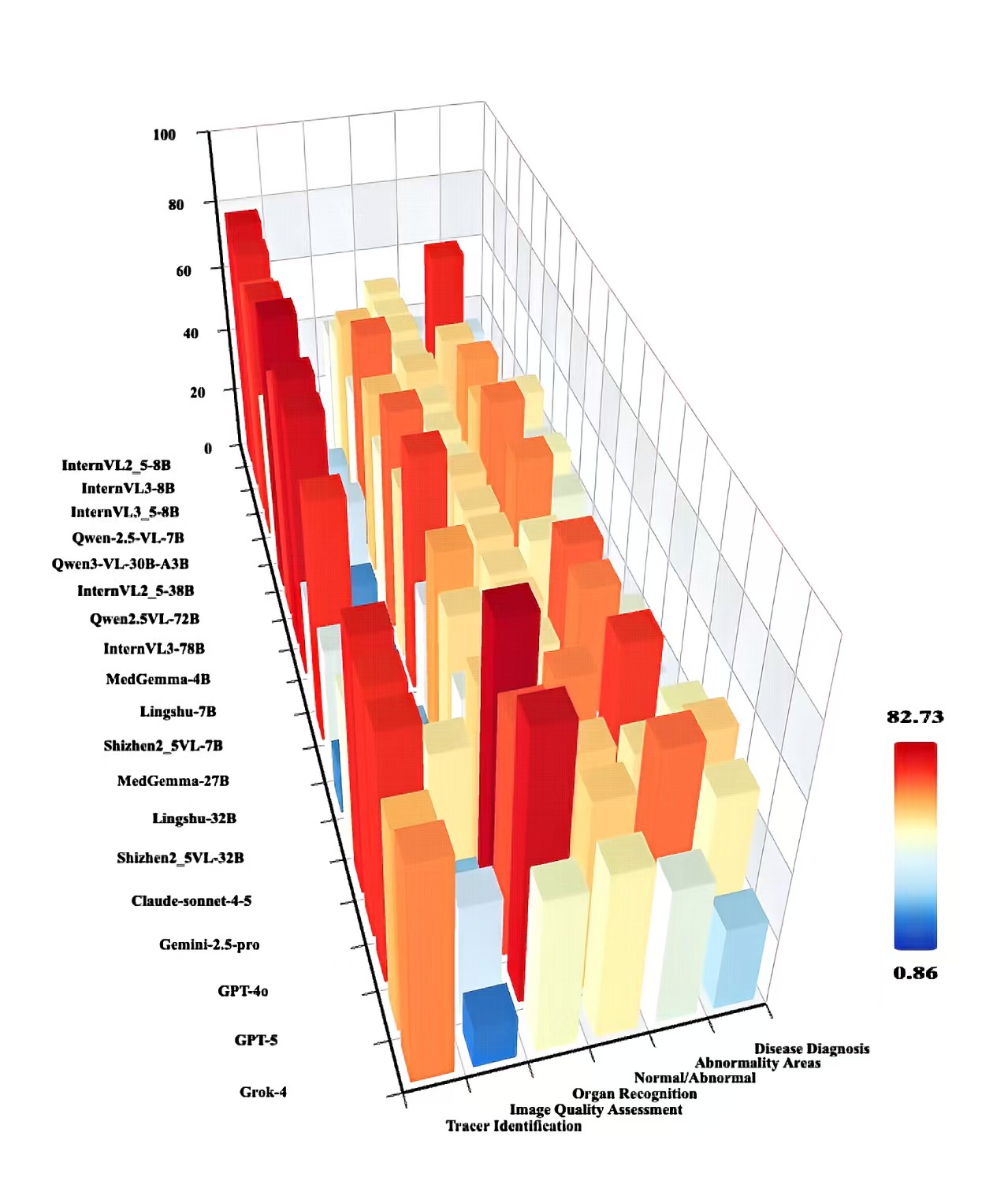} 
    \caption{\textbf{3D Visualization of Zero-Shot Performance across PET-Bench Tasks.} The height of each bar represents the accuracy of a specific model on a specific task. Models are ordered by family on the y-axis, and tasks are ordered hierarchically on the x-axis. Note: This figure provides an alternative visualization of the exact numerical results presented in Table~\ref{tab:zero_shot} of the main paper; no new experimental data is introduced.}
    \label{fig:sup_results}
\end{figure}

Fig.~\ref{fig:sup_results} visualizes the accuracy distribution across the six hierarchical tasks for all 19 MLLMs. It is important to note that this figure utilizes the identical data source as Table~\ref{tab:zero_shot} (Main Manuscript) and serves purely as a graphical supplement. This visualization highlights the performance stratification between generalist and medical-specific models across different task levels.

\section{Qualitative Case Gallery}
\label{sec:supp_cases}

To provide a concrete understanding of the PET-Bench tasks, we present representative examples for each of the five hierarchical levels.

\begin{figure}[H]
    \centering
    \begin{subfigure}[b]{0.32\linewidth}
        \includegraphics[width=\linewidth]{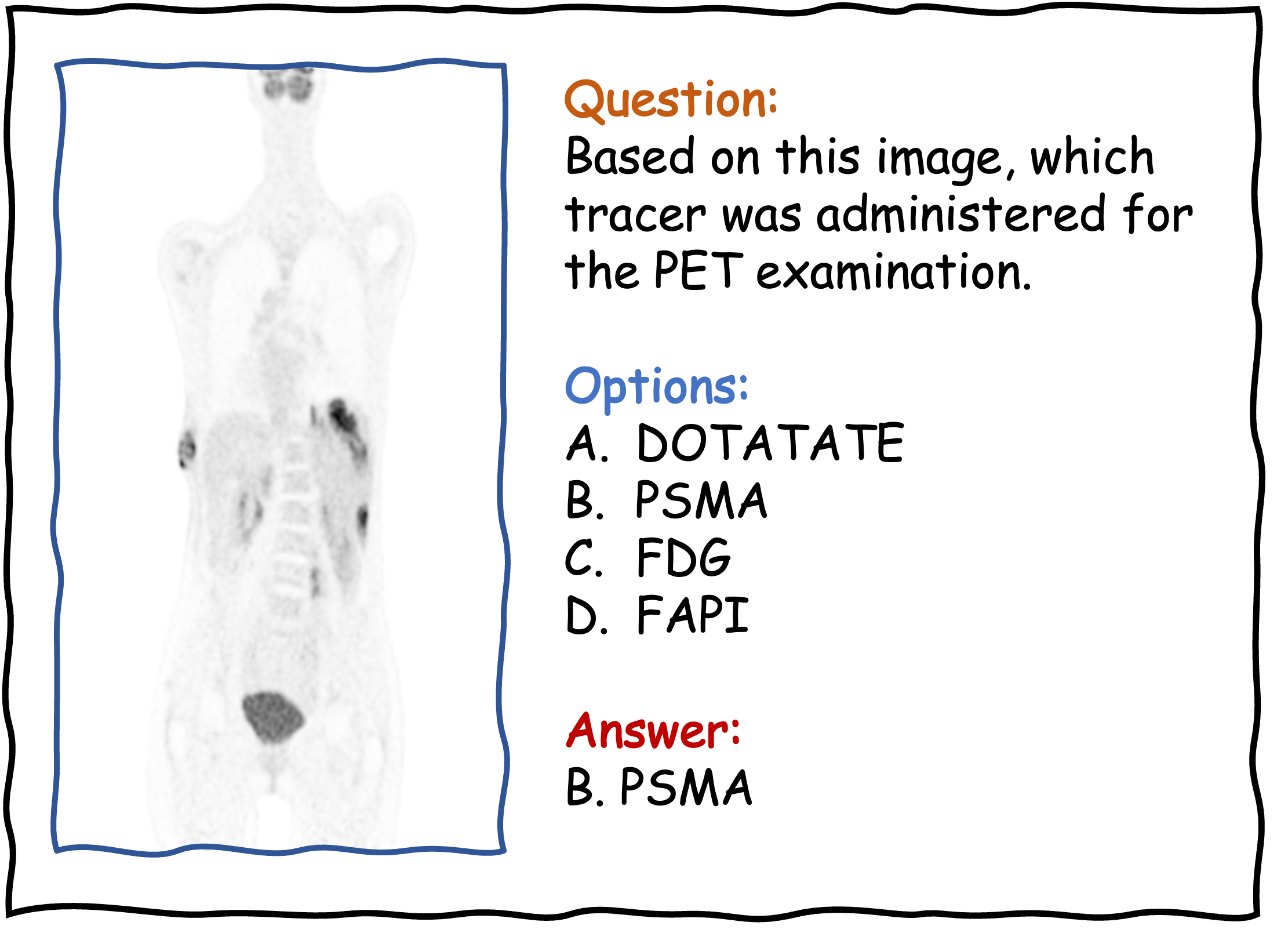}
        \caption{Case 1: PSMA}
    \end{subfigure}
    \begin{subfigure}[b]{0.32\linewidth}
        \includegraphics[width=\linewidth]{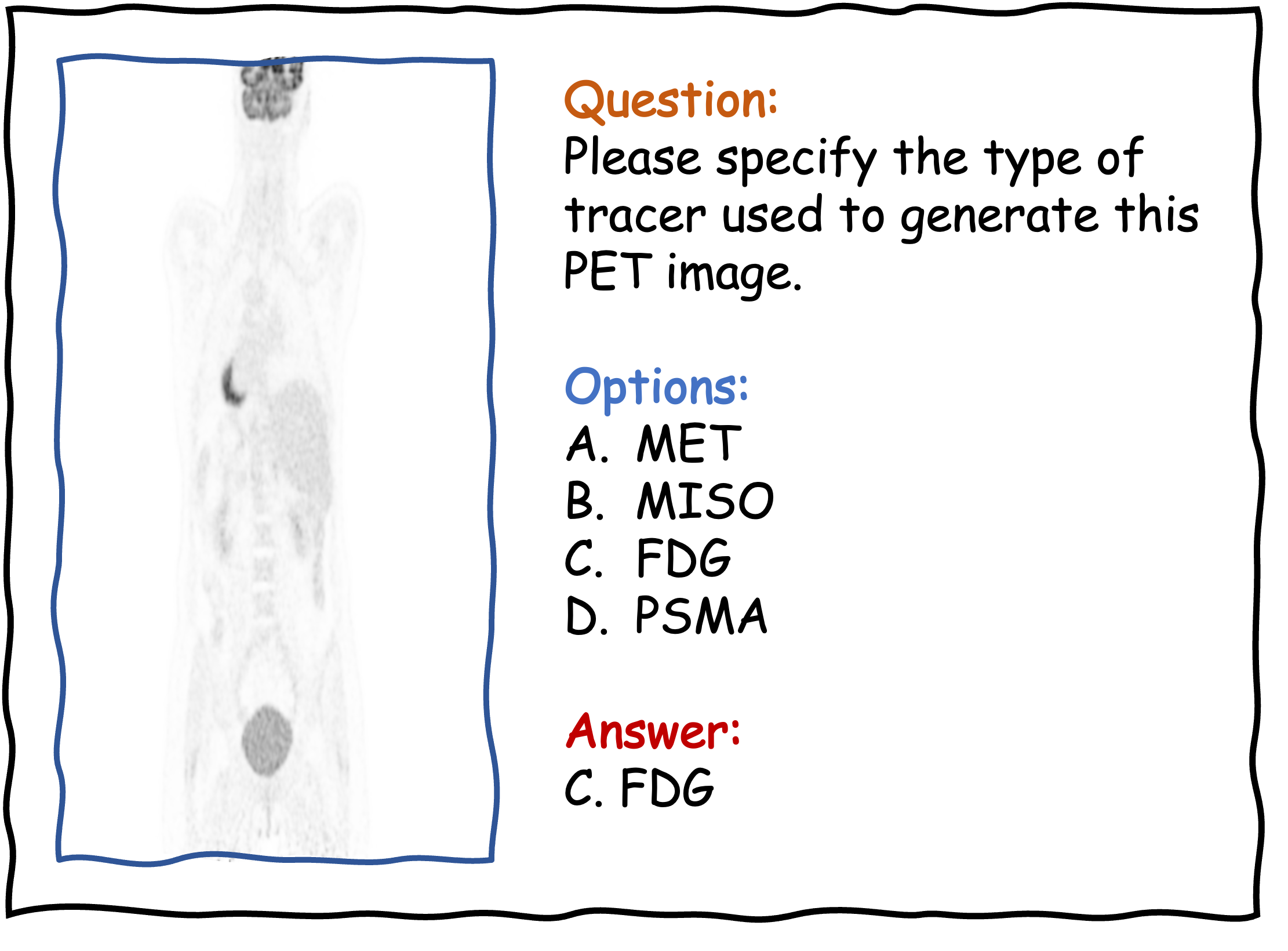}
        \caption{Case 2: FDG}
    \end{subfigure}
    \begin{subfigure}[b]{0.32\linewidth}
        \includegraphics[width=\linewidth]{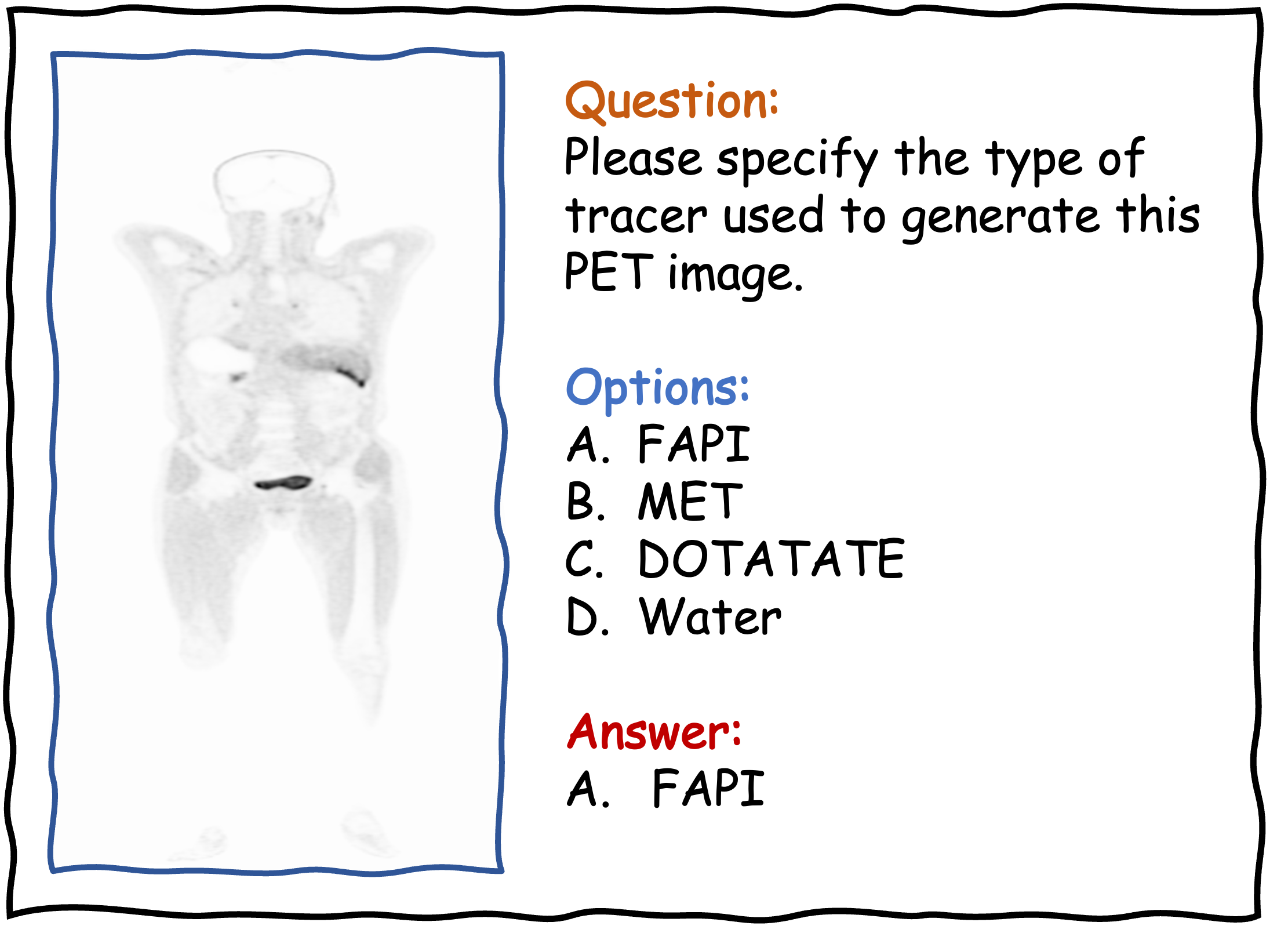}
        \caption{Case 3: FAPI}
    \end{subfigure}
    \caption{\textbf{Level 1: Tracer Identification.} Examples of different radiotracers. The model must identify the tracer (e.g., FDG, PSMA, FAPI) based on distinct physiological biodistribution patterns (e.g., brain uptake in FDG vs. salivary glands in PSMA).}
    \label{fig:supp_level1}
\end{figure}

\begin{figure}[H]
    \centering
    \begin{subfigure}[b]{0.32\linewidth}
        \includegraphics[width=\linewidth]{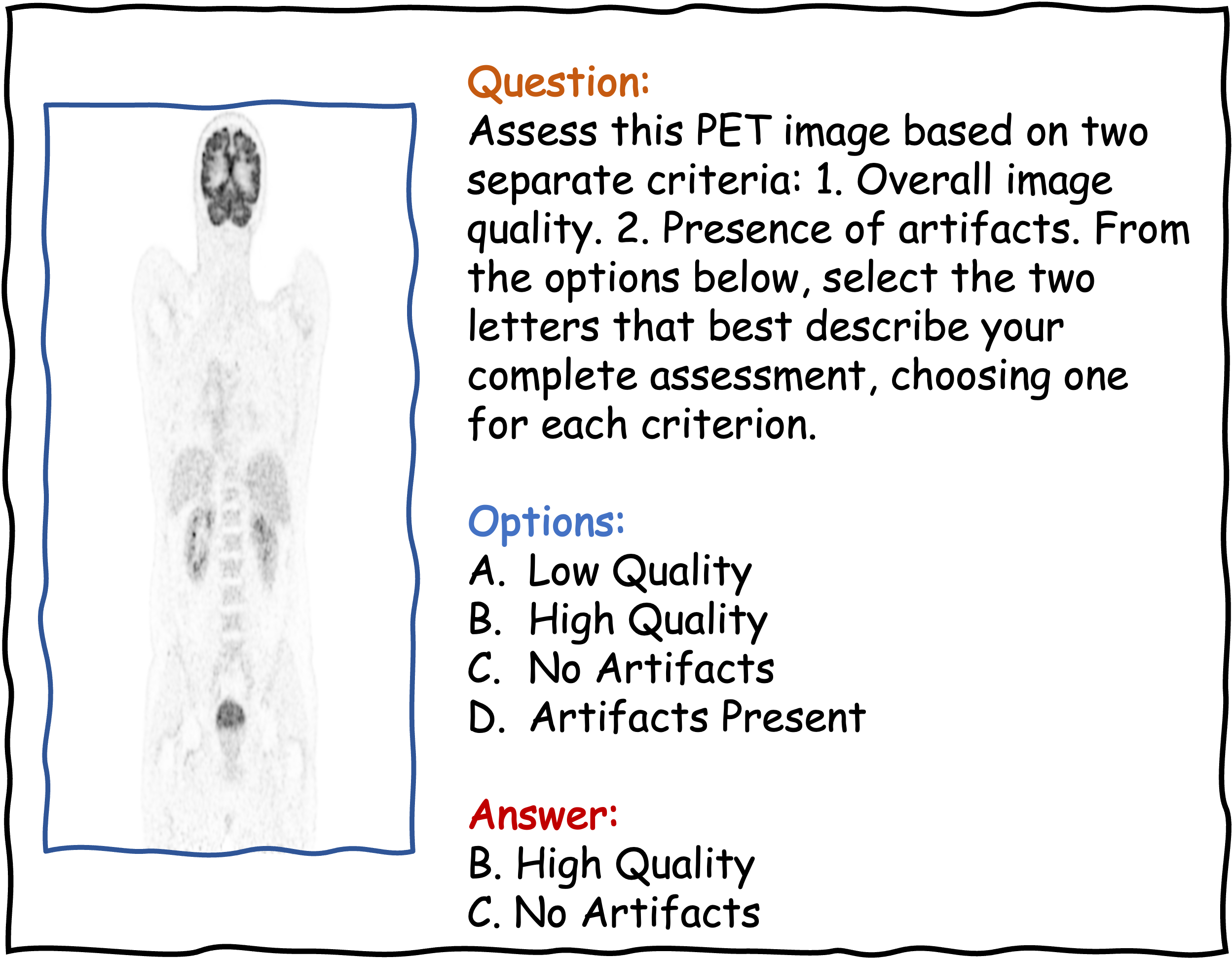}
        \caption{High Quality}
    \end{subfigure}
    \begin{subfigure}[b]{0.32\linewidth}
        \includegraphics[width=\linewidth]{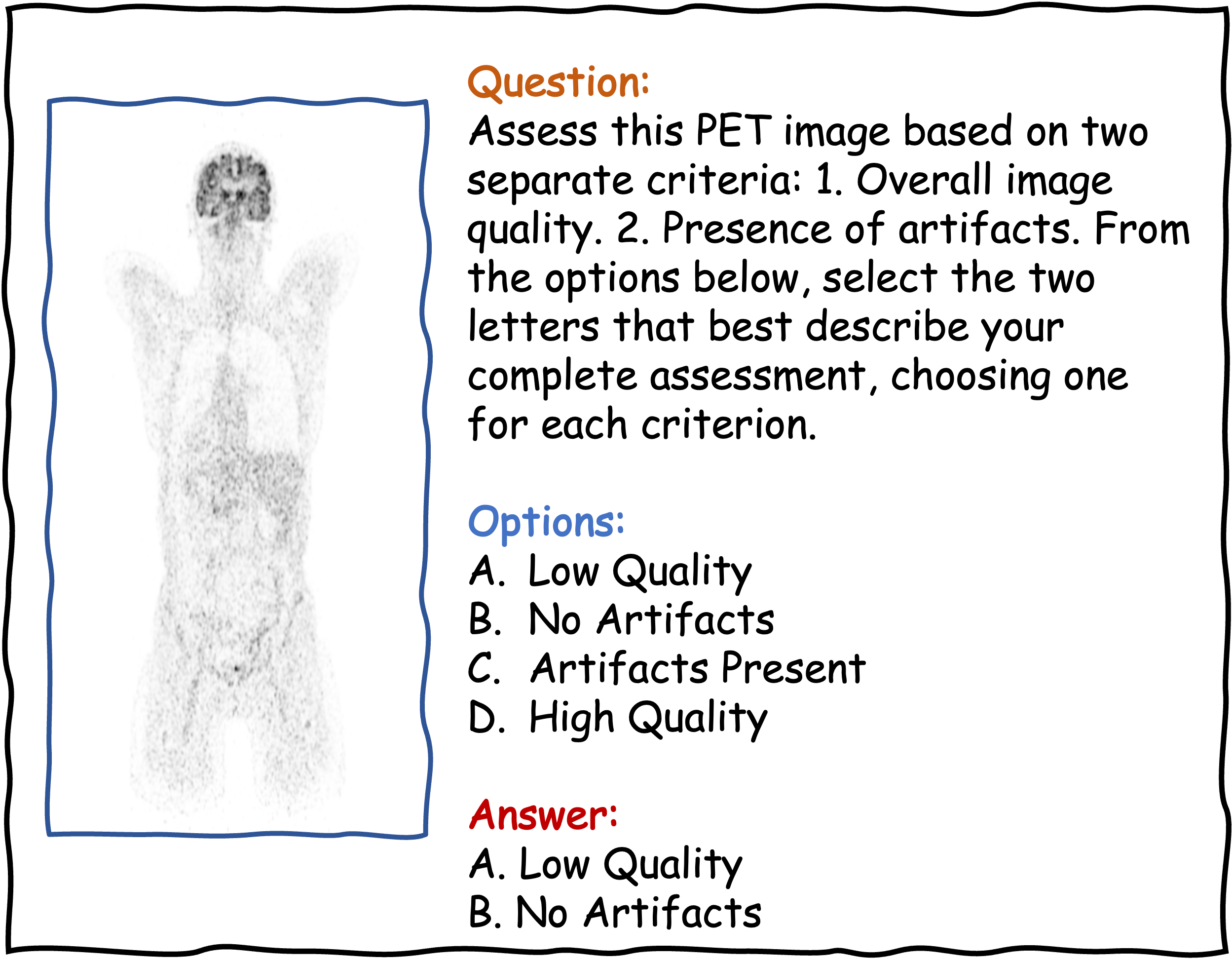}
        \caption{Low Quality (Low Dose)}
    \end{subfigure}
    \begin{subfigure}[b]{0.32\linewidth}
        \includegraphics[width=\linewidth]{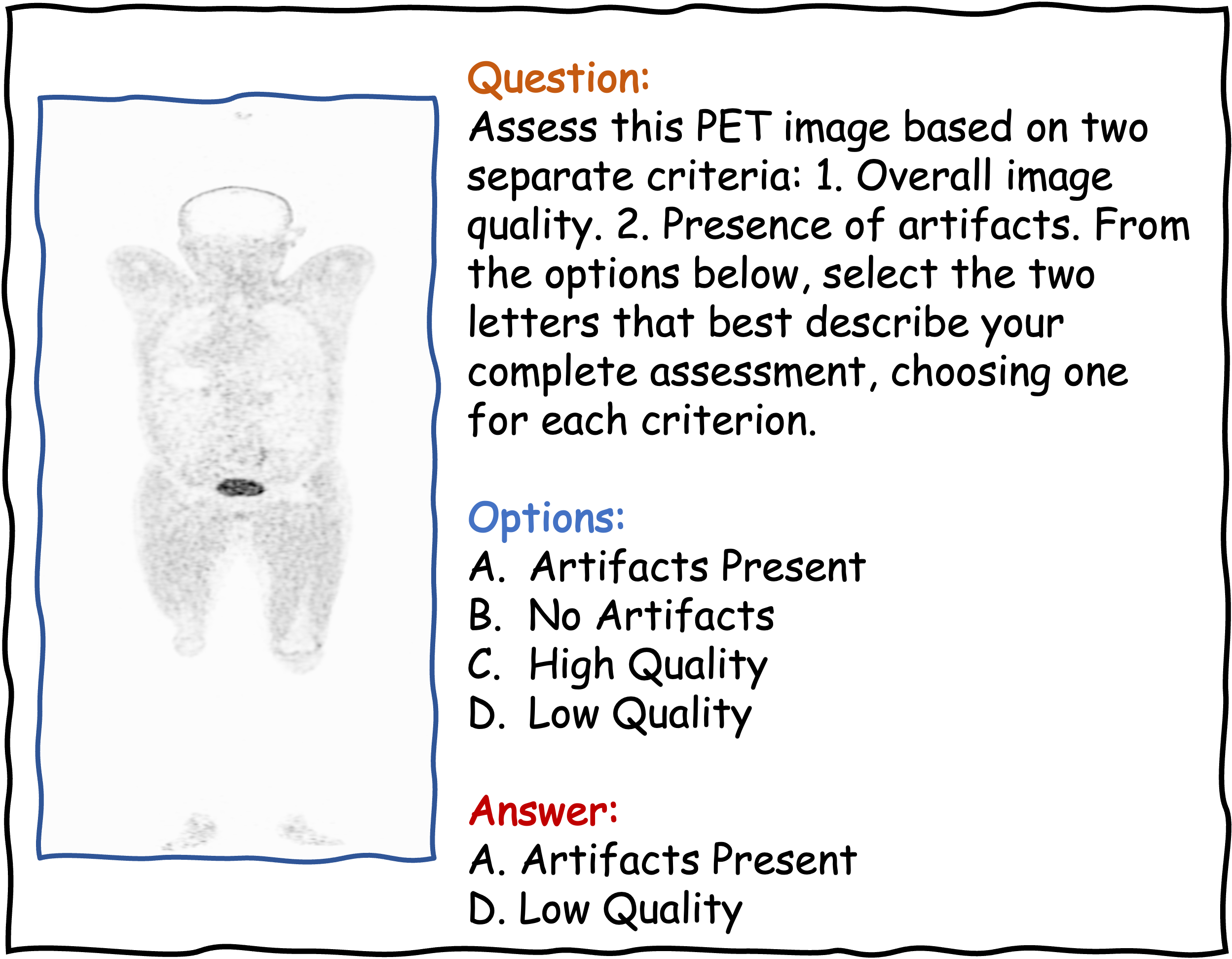}
        \caption{Artifact Presence}
    \end{subfigure}
    \caption{\textbf{Level 2: Image Quality Assessment.} Comparison of scans with varying quality. The task requires detecting noise degradation (center) or reconstruction artifacts (right) that compromise diagnostic utility.}
    \label{fig:supp_level2}
\end{figure}

\begin{figure}[H]
    \centering
    \begin{subfigure}[b]{0.32\linewidth}
        \includegraphics[width=\linewidth]{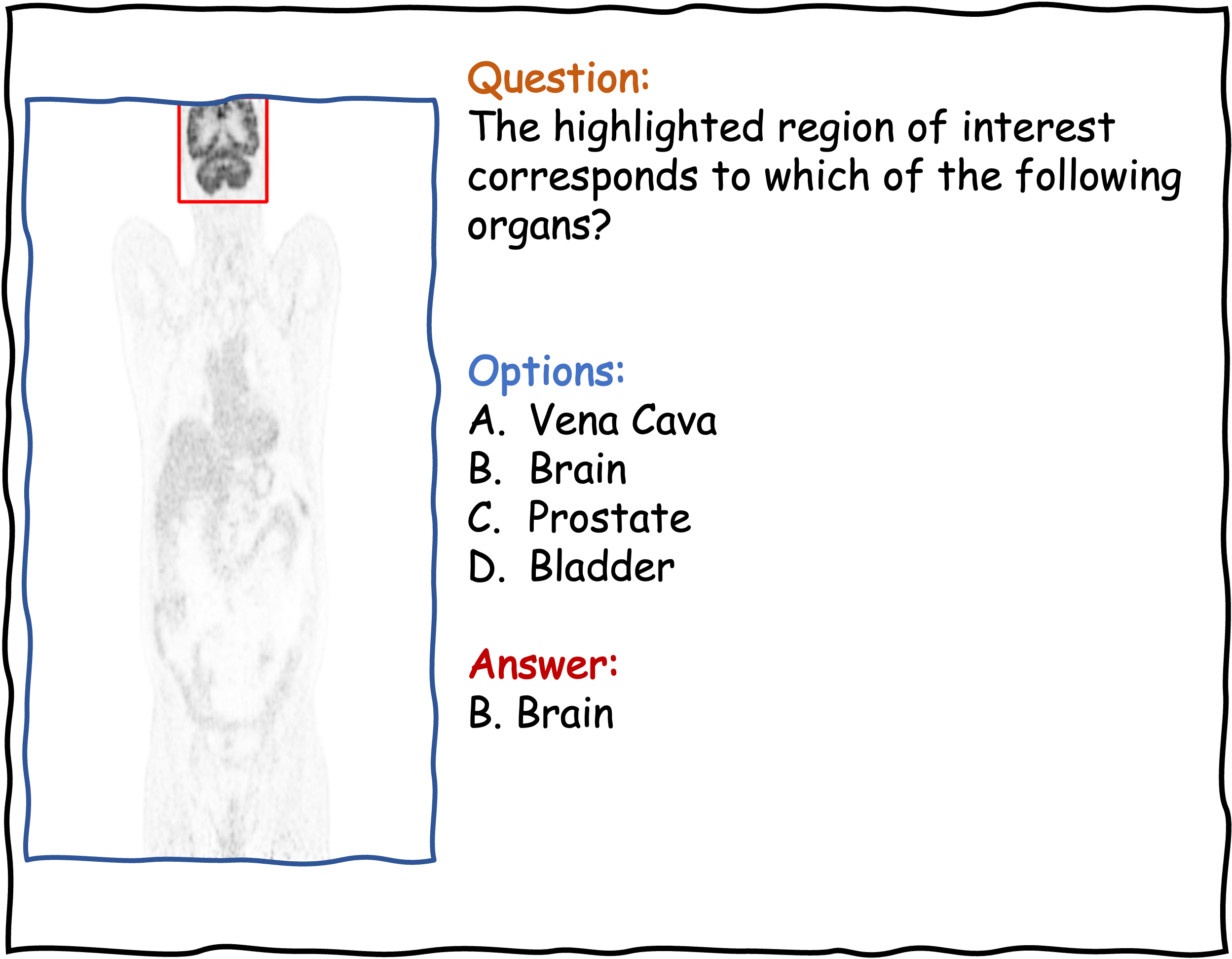}
        \caption{Case 1: Brain}
    \end{subfigure}
    \begin{subfigure}[b]{0.32\linewidth}
        \includegraphics[width=\linewidth]{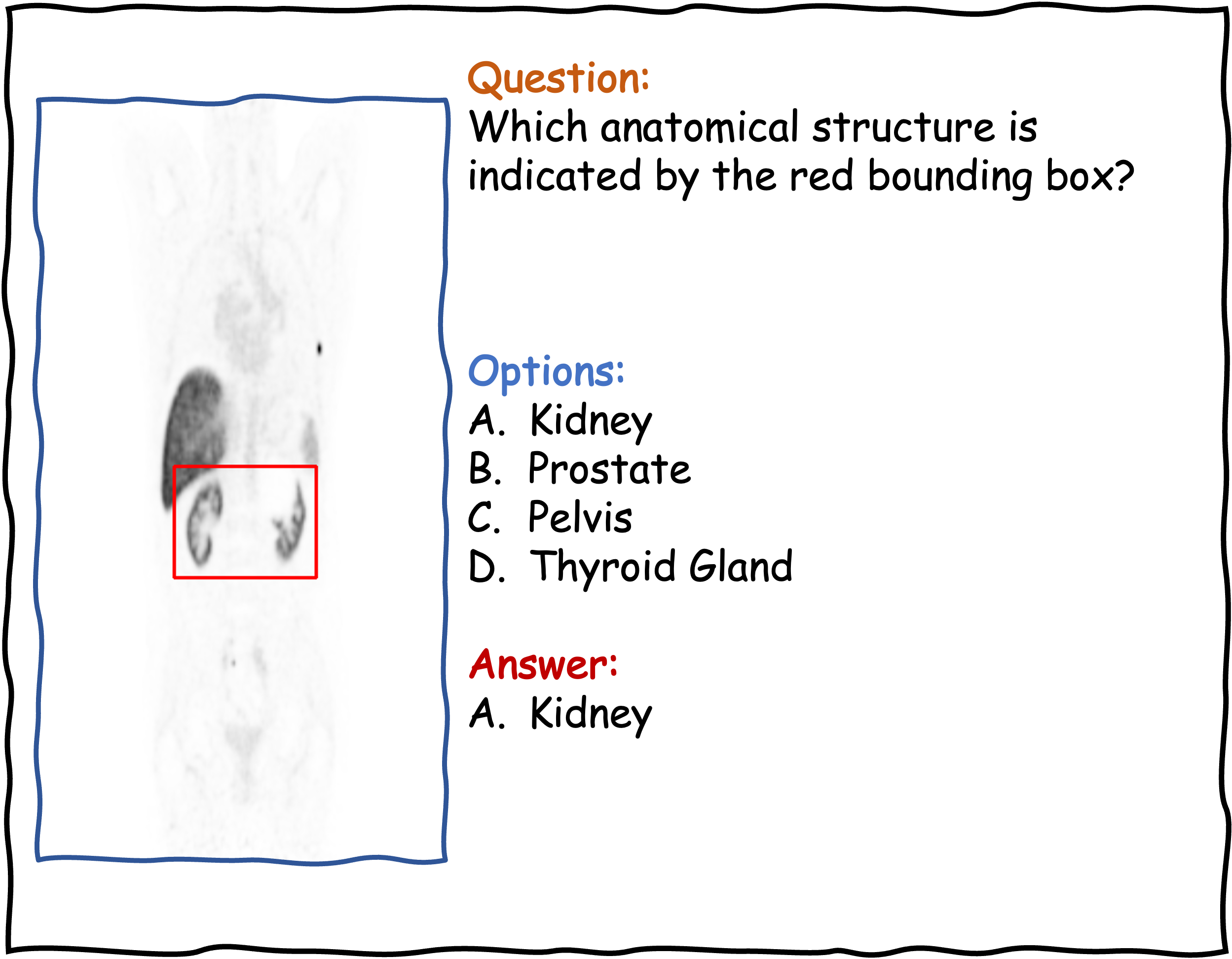}
        \caption{Case 2: Kidney}
    \end{subfigure}
    \begin{subfigure}[b]{0.32\linewidth}
        \includegraphics[width=\linewidth]{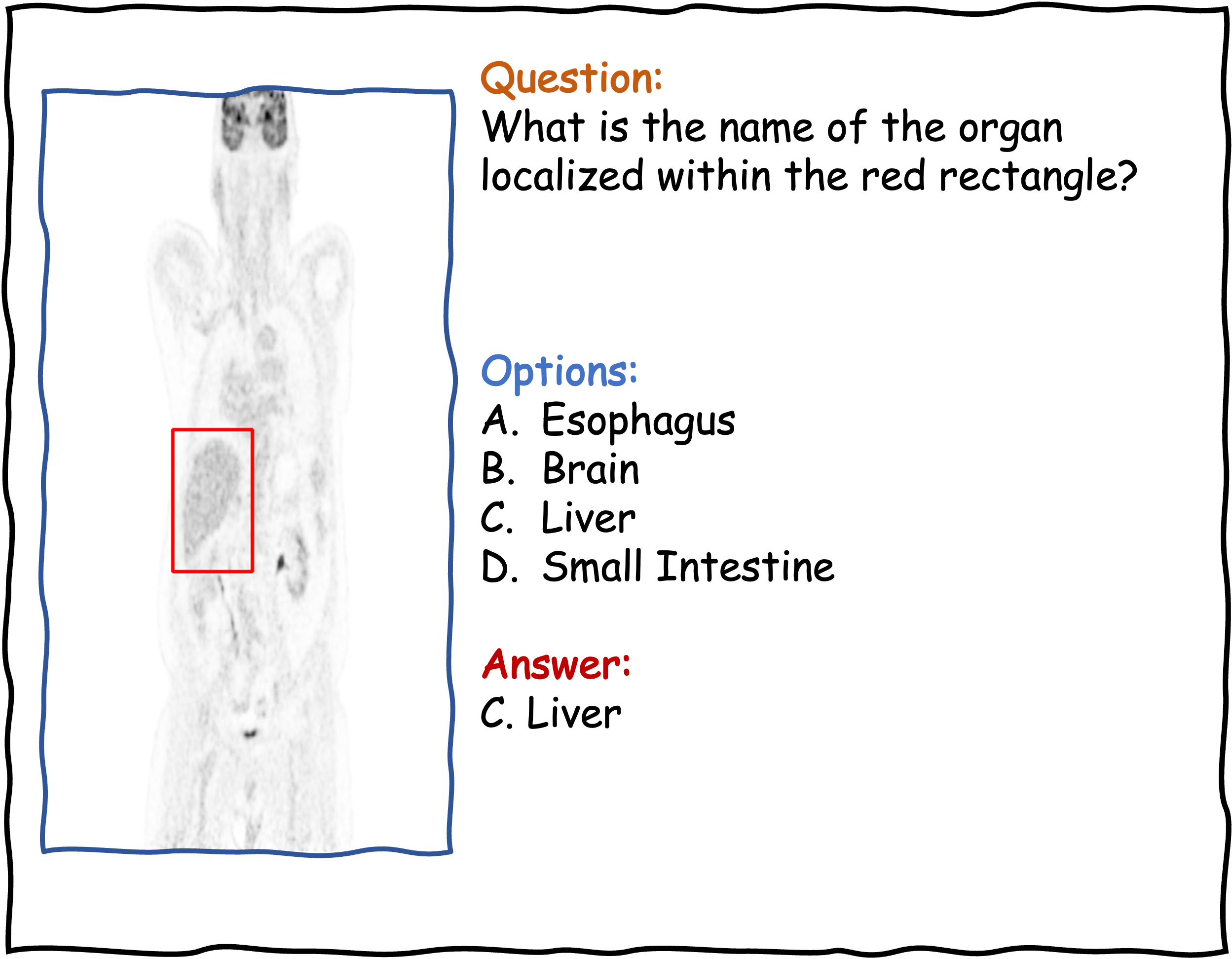}
        \caption{Case 3: Liver}
    \end{subfigure}
    \caption{\textbf{Level 3: Organ Recognition.} The model must identify the anatomical structure indicated by the bounding box based solely on metabolic intensity and shape, without CT anatomical guidance.}
    \label{fig:supp_level3}
\end{figure}

\begin{figure}[H]
    \centering
    \begin{subfigure}[b]{0.32\linewidth}
        \includegraphics[width=\linewidth]{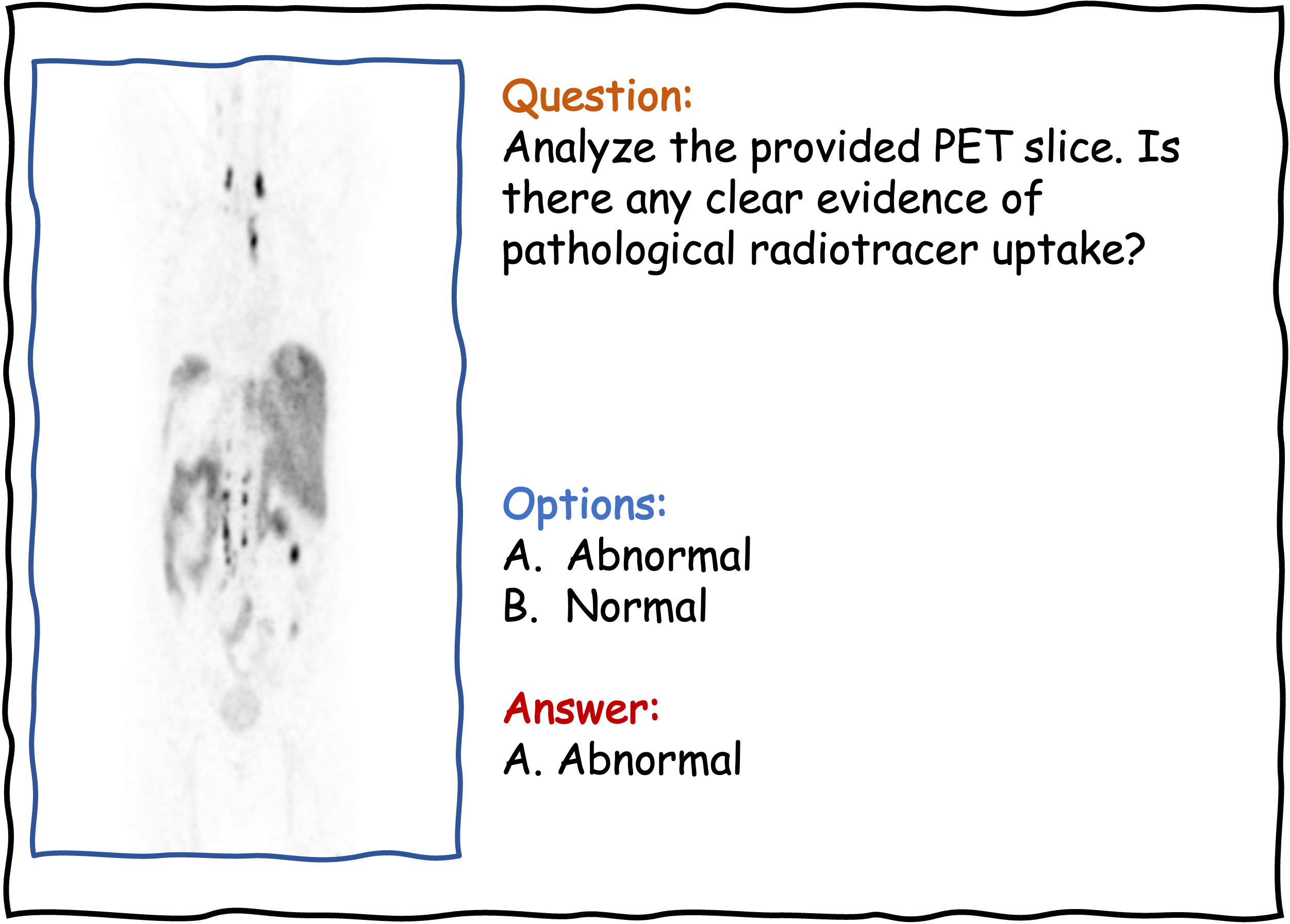}
        \caption{Abnormal 1}
    \end{subfigure}
    \begin{subfigure}[b]{0.32\linewidth}
        \includegraphics[width=\linewidth]{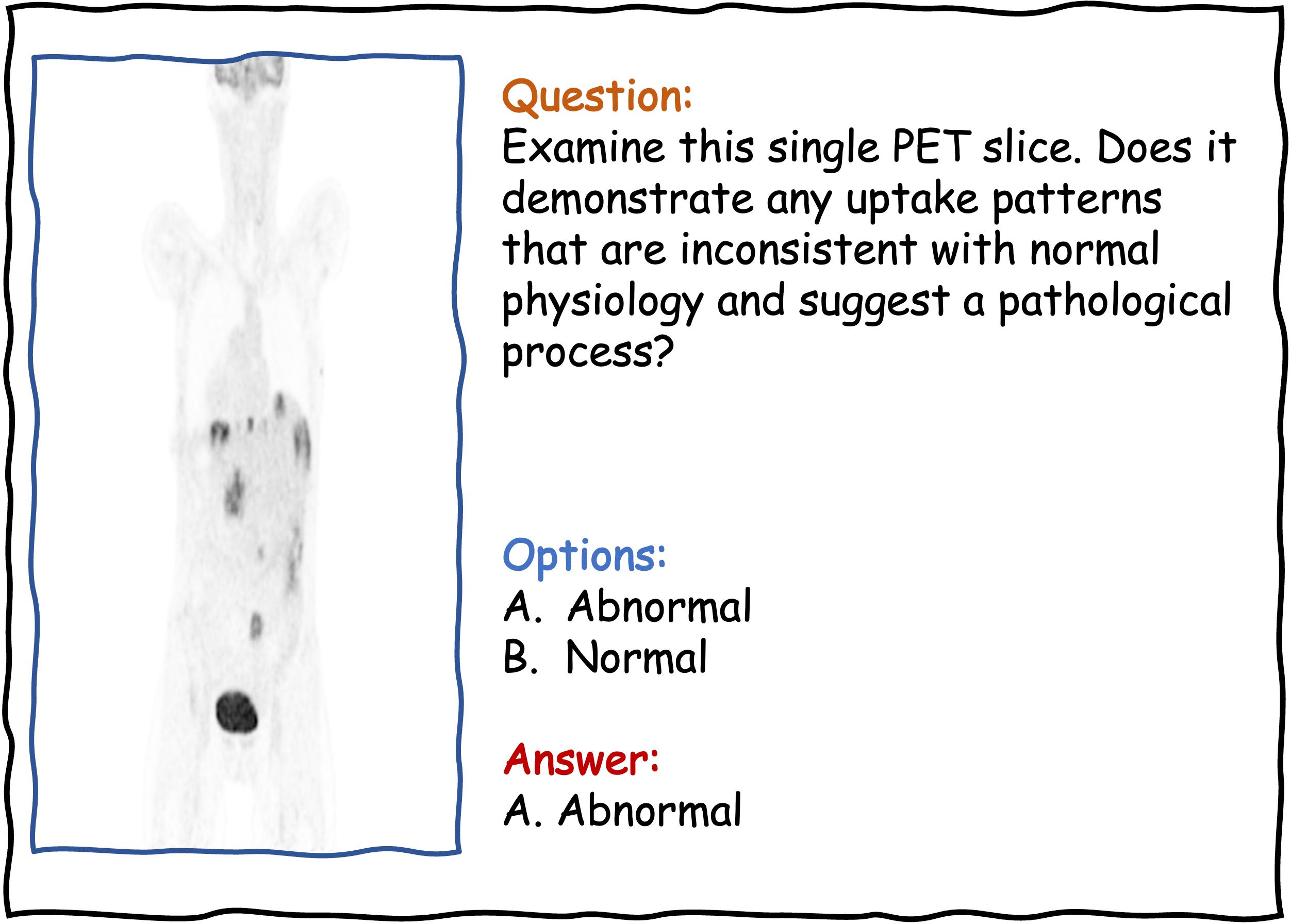}
        \caption{Abnormal 2}
    \end{subfigure}
    \begin{subfigure}[b]{0.32\linewidth}
        \includegraphics[width=\linewidth]{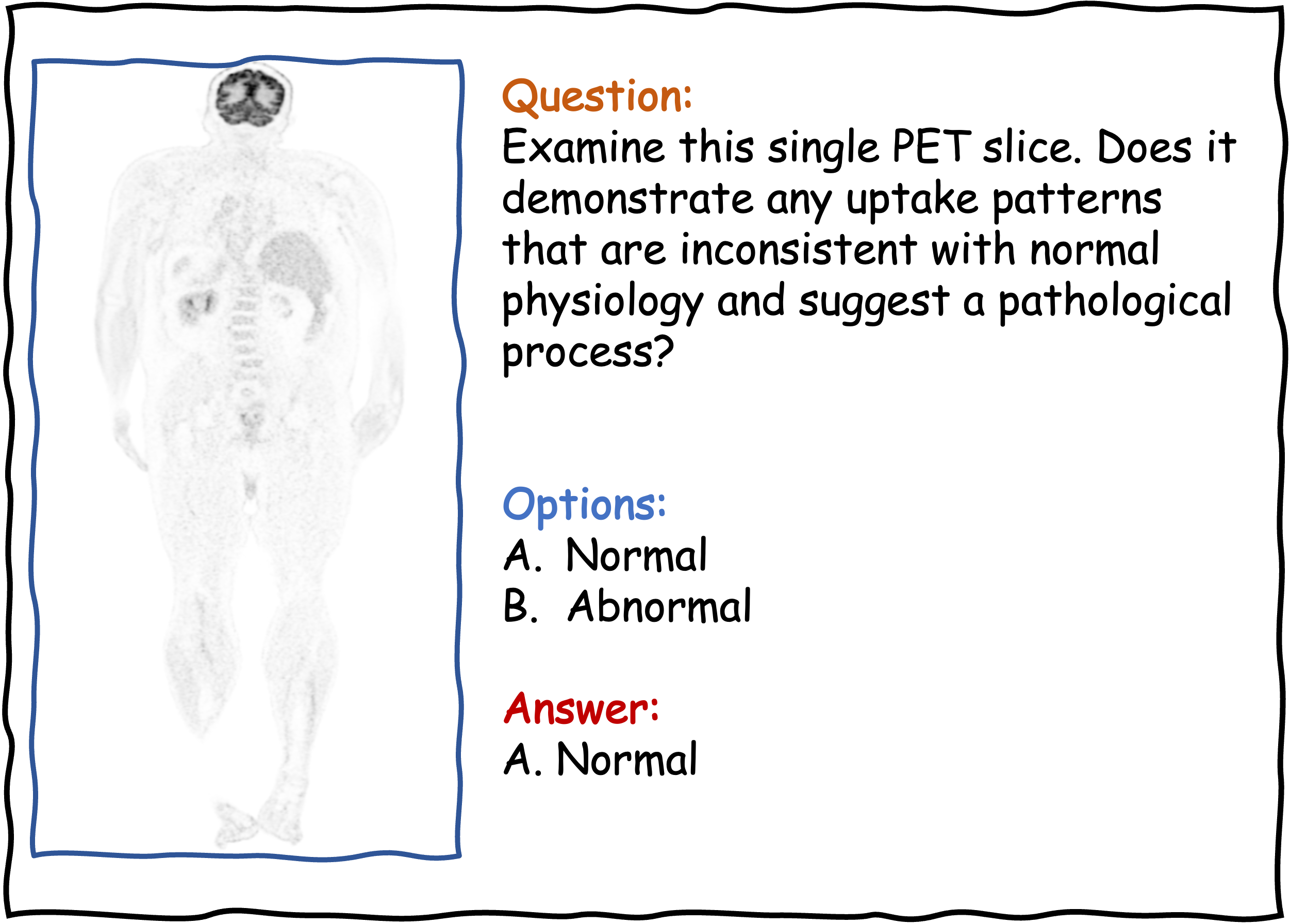}
        \caption{Normal}
    \end{subfigure}
    \caption{\textbf{Level 4a: Abnormality Identification.} Distinguishing between pathological uptake (a, b) and normal physiological background (c).}
    \label{fig:supp_level4a}
\end{figure}

\begin{figure}[H]
    \centering
    \begin{subfigure}[b]{0.32\linewidth}
        \includegraphics[width=\linewidth]{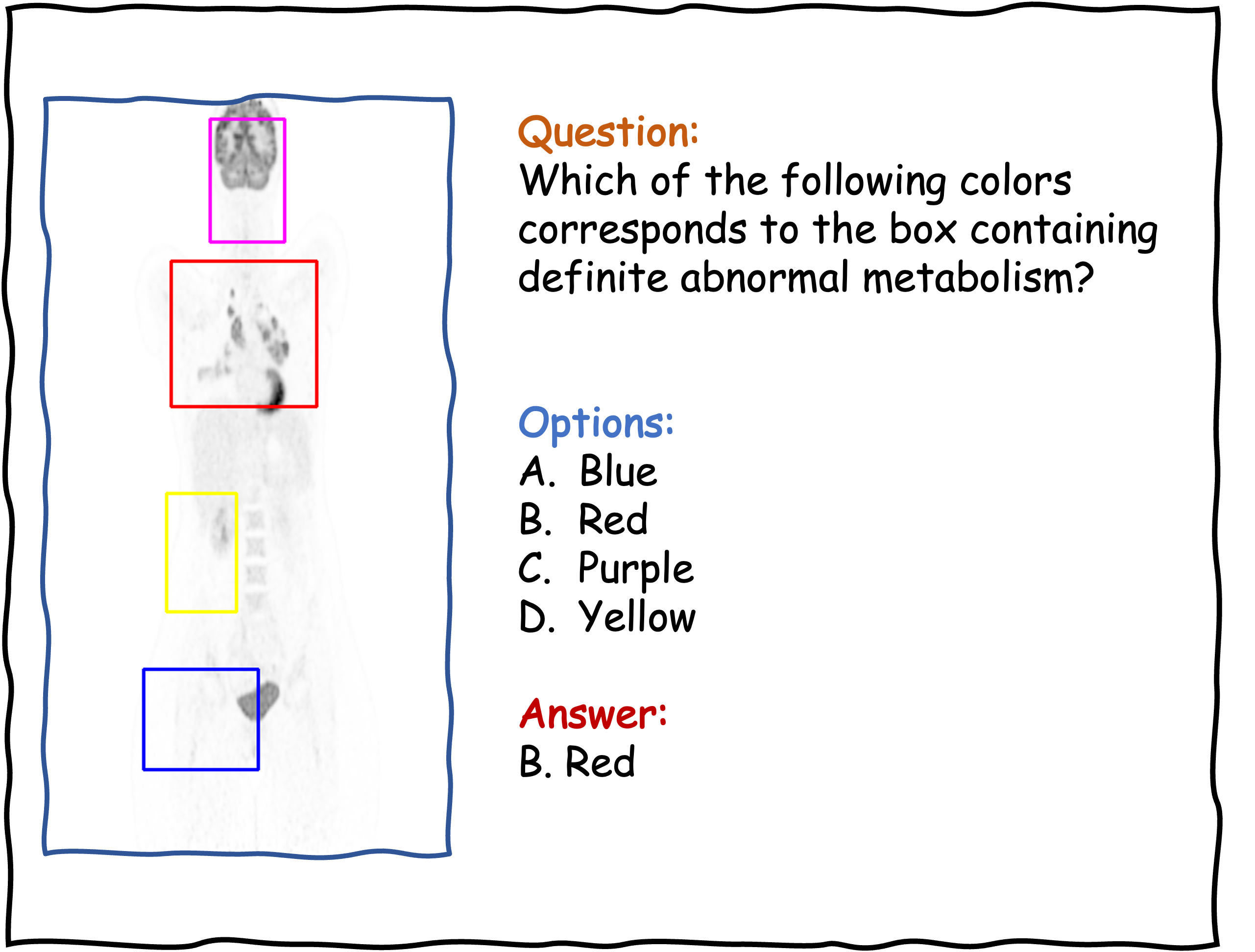}
        \caption{Localization 1}
    \end{subfigure}
    \begin{subfigure}[b]{0.32\linewidth}
        \includegraphics[width=\linewidth]{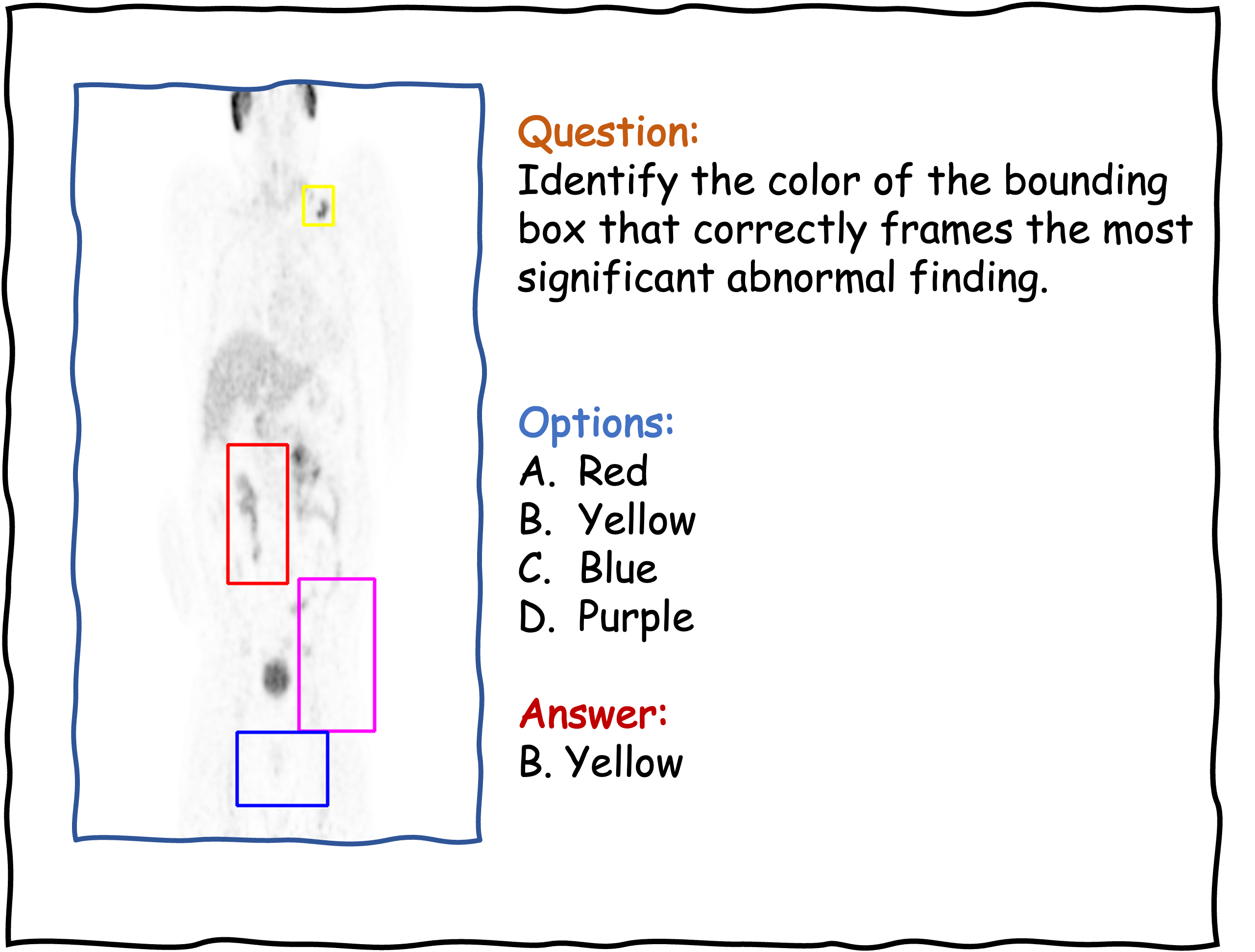}
        \caption{Localization 2}
    \end{subfigure}
    \begin{subfigure}[b]{0.32\linewidth}
        \includegraphics[width=\linewidth]{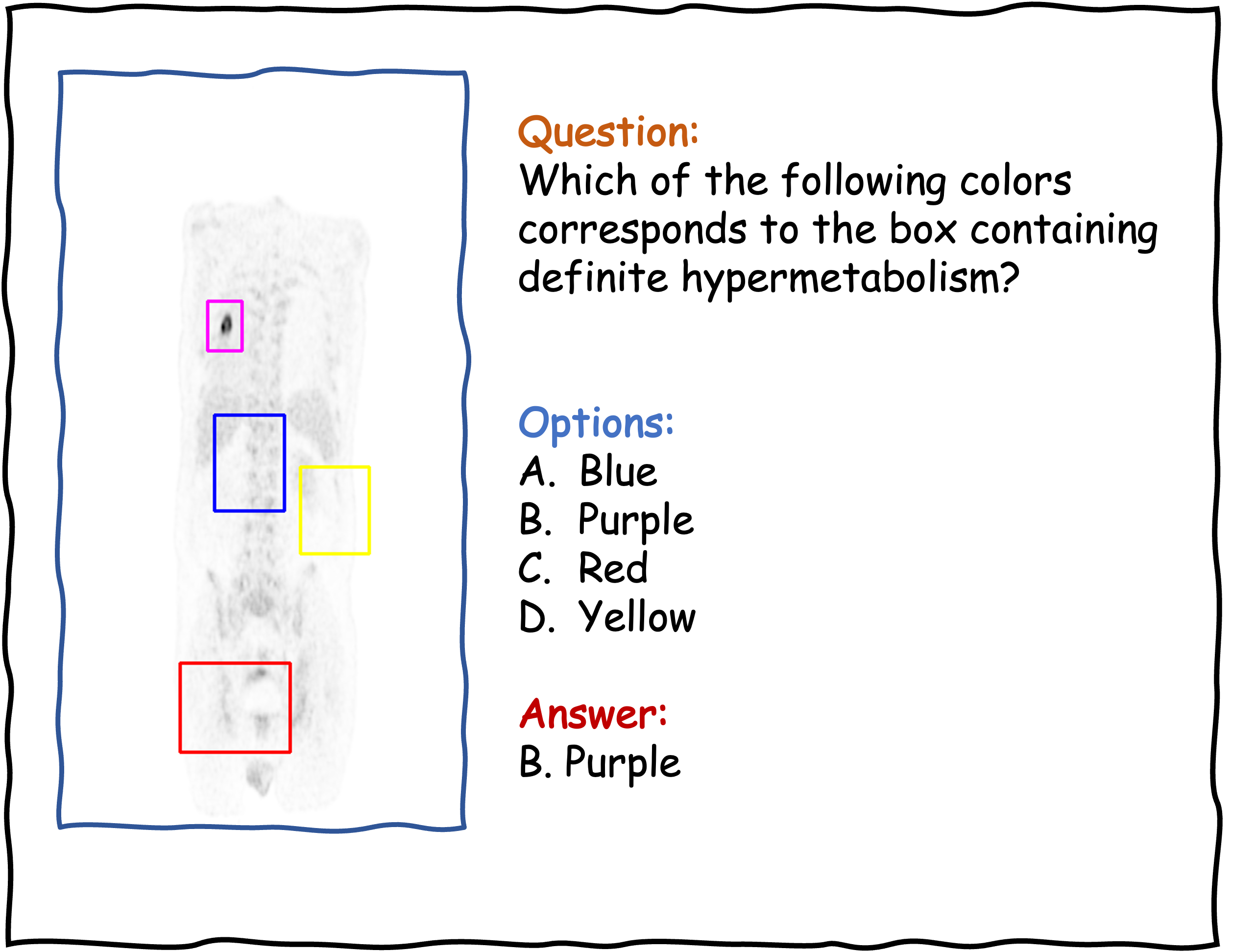}
        \caption{Localization 3}
    \end{subfigure}
    \caption{\textbf{Level 4b: Abnormality Localization.} The model must identify the specific region (color-coded box) containing hypermetabolic lesions.}
    \label{fig:supp_level4b}
\end{figure}

\begin{figure}[H]
    \centering
    \includegraphics[width=0.9\linewidth]{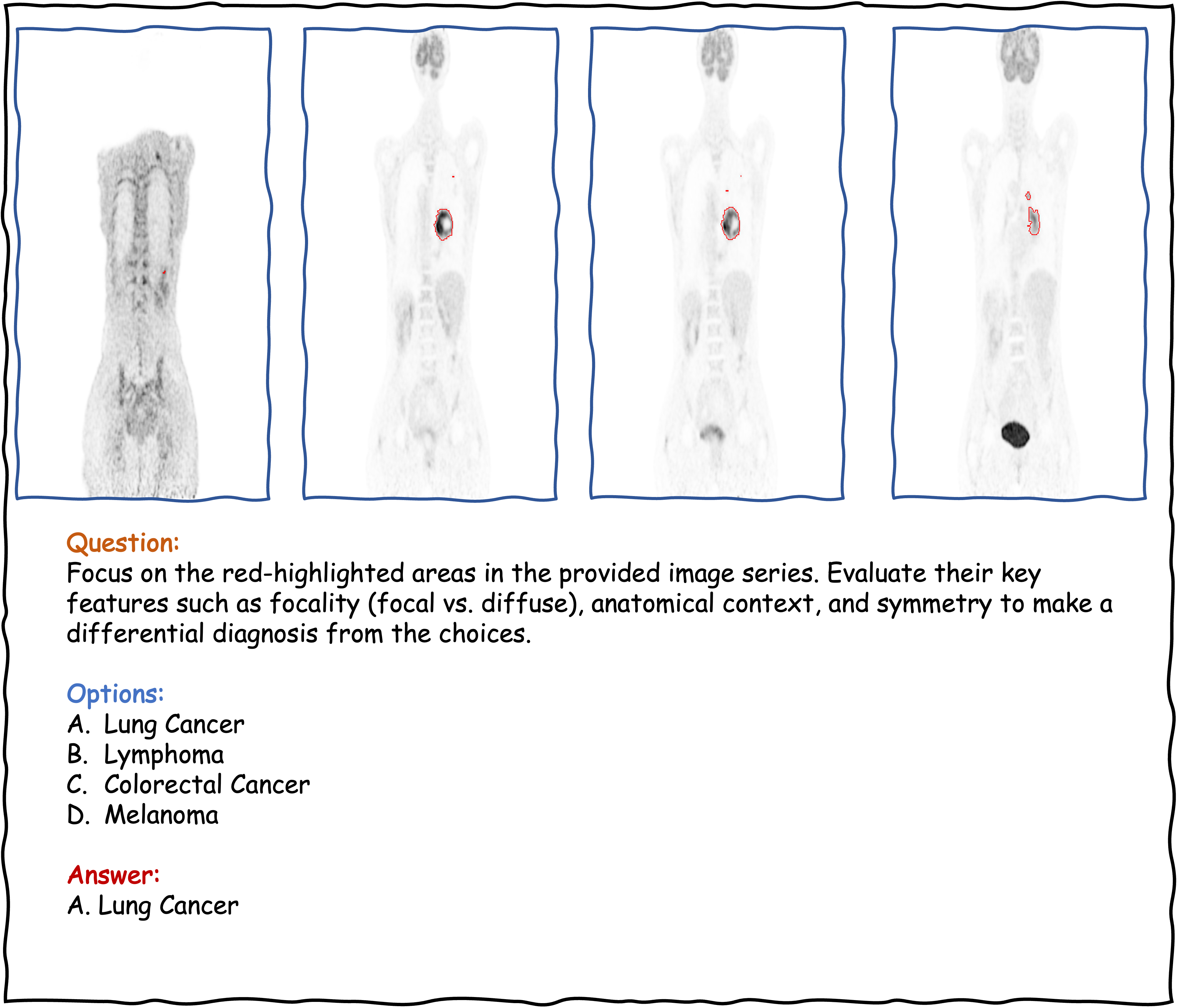}
    \caption{\textbf{Level 5: Disease Diagnosis (Case 1).} Multi-slice input showing disease progression. The model must synthesize findings to predict the specific pathology (e.g., Lung Cancer).}
    \label{fig:supp_level5_1}
\end{figure}

\begin{figure}[H]
    \centering
    \includegraphics[width=0.9\linewidth]{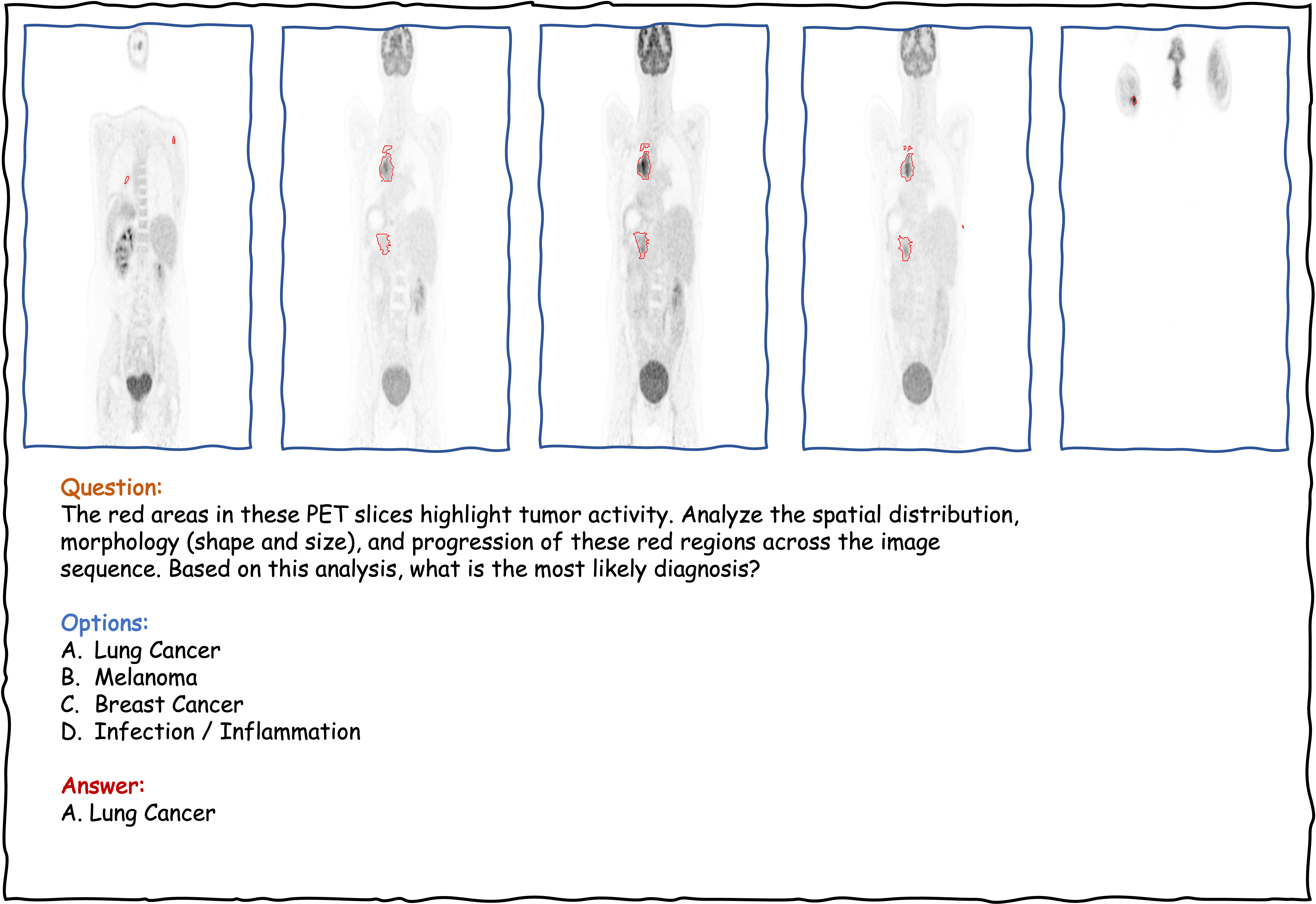}
    \caption{\textbf{Level 5: Disease Diagnosis (Case 2).} Complex case requiring differentiation between lymphoma and sarcoidosis based on distribution patterns.}
    \label{fig:supp_level5_2}
\end{figure}

\begin{figure}[H]
    \centering
    \includegraphics[width=0.9\linewidth]{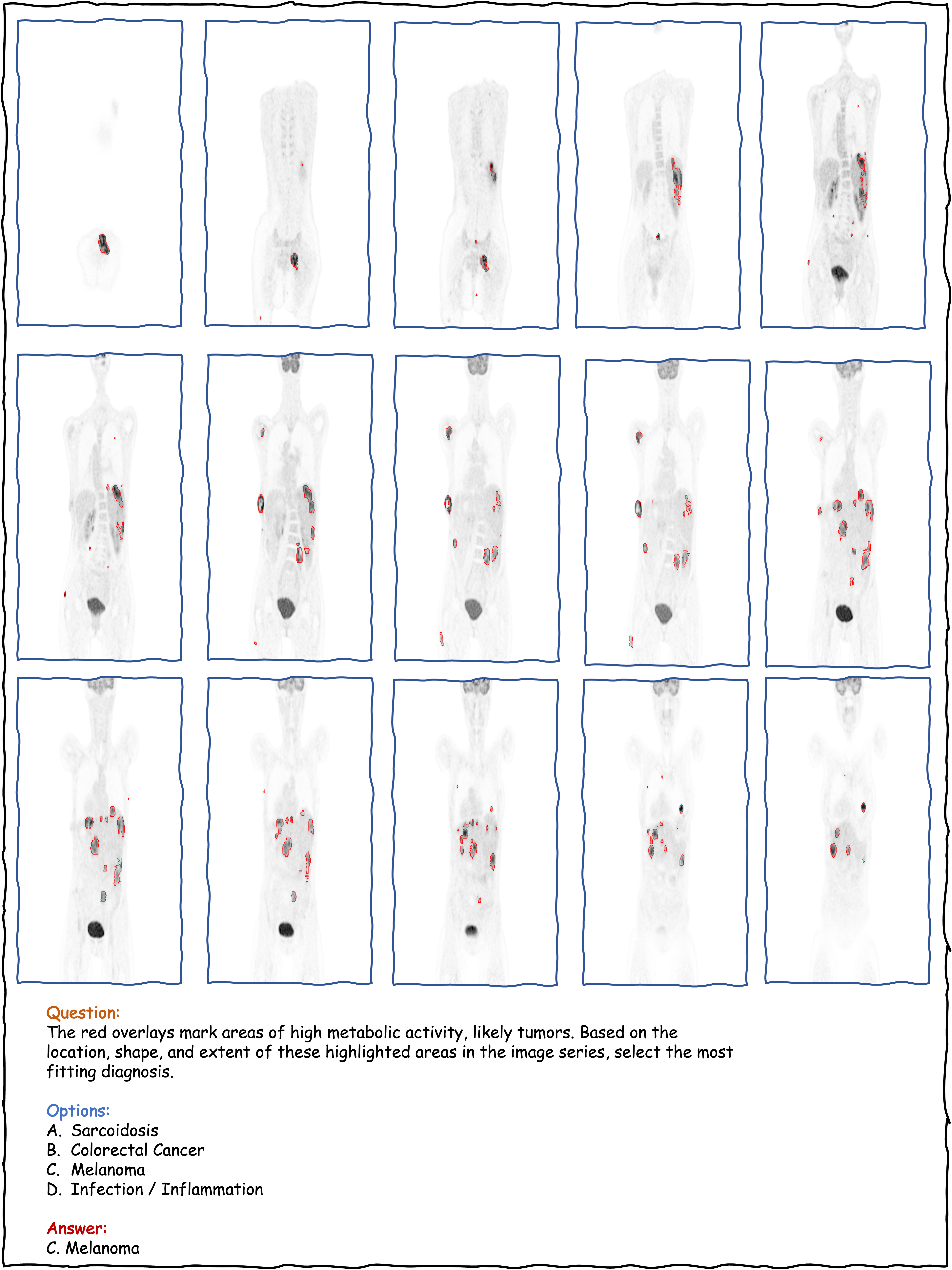}
    \caption{\textbf{Level 5: Disease Diagnosis (Case 3).} Diagnosis of metastatic disease requiring whole-body assessment.}
    \label{fig:supp_level5_3}
\end{figure}

\end{document}